\documentclass[11pt, a4paper]{article}
\usepackage{fullpage}
\usepackage{amsfonts}
\usepackage{amssymb}
\usepackage{amsmath}
\usepackage{amsthm}
\usepackage{graphicx}
\usepackage[makeroom]{cancel}
\usepackage{enumitem}
\usepackage{url}
\usepackage[margin=.9in]{geometry}
\usepackage{amsopn}
\usepackage{mathtools}
\usepackage{hyperref}
\usepackage{doi}
\usepackage{cite}
\usepackage{bbm}
\usepackage{xcolor}
\usepackage[symbol]{footmisc}
\usepackage{algorithm}
\usepackage{algpseudocode}
\usepackage{booktabs}
\usepackage{caption}
\usepackage{multirow}
\usepackage{makecell}
\usepackage{subcaption}
\theoremstyle{definition}

\DeclareMathOperator*{\argmin}{arg\,min}
\newcommand{\R}{\mathbb{R}}
\newcommand{\C}{\mathbb{C}}

\newcommand{\w}{\boldsymbol{\omega}}
\newcommand{\Phib}{\mathbf{\Phi}_\mathbf{b}}
\newcommand{\Phibhat}{\hat{\mathbf{\Phi}}_\mathbf{b}}
\newcommand{\Tw}{\mathbf{T}(\boldsymbol{\omega})}

\begin{document}
\begin{center}
    \Large \bf Sparse-mode Dynamic Mode Decomposition for Disambiguating Local and Global Structures
\end{center}
\begin{center}
    Sara M. Ichinaga$^{1}$\footnote[1]{Corresponding authors (sarami7@uw.edu)}, Steven L. Brunton$^{2}$, Aleksandr Y. Aravkin$^{1}$, J. Nathan Kutz$^{1,3}$
\end{center}
\begin{center}
    \scriptsize{
    ${}^1$ Department of Applied Mathematics, University of Washington, Seattle, WA 98195, United States \\ 
    ${}^2$ Department of Mechanical Engineering, University of Washington, Seattle, WA 98195, United States \\
    ${}^3$ Department of Electrical and Computer Engineering, University of Washington, Seattle, WA 98195, United States
    }
\end{center}

\begin{abstract}
The dynamic mode decomposition (DMD) is a data-driven approach that extracts the dominant features from spatiotemporal data. In this work, we introduce sparse-mode DMD, a new variant of the optimized DMD framework that specifically leverages sparsity-promoting regularization in order to approximate DMD modes which have localized spatial structure. The algorithm maintains the noise-robust properties of optimized DMD while disambiguating between modes which are spatially local versus global in nature.  In many applications, such modes are associated with discrete and continuous spectra respectively, thus allowing the algorithm to explicitly construct, in an unsupervised manner, the distinct portions of the spectrum. We demonstrate this by analyzing synthetic and real-world systems, including examples from optical waveguides, quantum mechanics, and sea surface temperature data.
\end{abstract}

\section{Introduction}
\label{sec:introduction}

Spatio-temporal data dominate many scientific disciplines, including climate and weather patterns, whole-brain activity, disease vectors, or mechanical systems subject to the laws of physics.  These systems are typically dynamical systems whose underlying governing equations are often unknown, or only partially known.  Regardless, they can be thought of as possessing an underlying governing equation of the form
\begin{equation}
    \dot{\mathbf{x}}(t) = f(\mathbf{x}(t); \boldsymbol{\mu}),
\end{equation}
where $\mathbf{x}(t) \in \C^n$ are snapshot observations, $t \in \R$ are points in time, $\boldsymbol{\mu} \in \R^\ell$ are parameters of the system, and the function $f$ defines a system of governing equations. 
Measurement noise, unknown or partially known $f$, and high computational costs compromise the process of building accurate and effective equation-free reduced order models. It is hence of great importance and scientific relevance to develop and build upon mathematical tools and algorithms that aim to robustly extract meaning from real-world time-varying data sets.

The \textit{dynamic mode decomposition} (DMD) \cite{schmid_2010, tu_2014, dmd_book, schmid2022dynamic} seeks to decompose time-varying data sets into a low-rank coherent set of spatiotemporal structures.  Specifically, the temporal evolution is assumed to be approximated by a linear dynamical system $\dot{\mathbf{x}}(t) = \mathbf{A} \mathbf{x}(t)$ whose evolution dynamics are exponentials of the form
\begin{equation}
    \mathbf{x}(t) = \sum_{j=1}^N b_j \boldsymbol{\phi}_j \exp(\omega_j t)
    \label{eq:DMDsolution}
\end{equation}
with the imaginary component of the exponent $\omega_j$ modeling periodic temporal oscillations.  Regressing to the solution form (\ref{eq:DMDsolution}) through optimized DMD~\cite{optdmd,bopdmd} has many benefits, and it allows for tasks such as model reduction and future-state prediction. This insight paired with the equation-free nature of the algorithm has led to the widespread application of DMD across many scientific fields, including fluid dynamics \cite{schmid_2008, schmid_2009, schmid_2010, Noack2016jfm}, epidemiology \cite{proctor_2015}, neuroscience \cite{brunton_2016, alfatlawi_2020}, finance \cite{mann_2016}, robotics \cite{berger_2015, abraham_2019, bruder_2019}, plasma physics \cite{taylor_2018, kaptanoglu_2020} and control~\cite{dmd_book} to name a few. The method has additionally been connected to Koopman operator theory and has since become the standard approach for approximating the Koopman operator from data \cite{rowley_2009, tu_2014, modern_koopman}.

Over the years since the algorithm's conception, the DMD algorithm has seen a great deal of improvements, variants, and extensions \cite{schmid2022dynamic}, with one of the most notable innovations being the optimized DMD algorithm \cite{optdmd}. Optimized DMD and its more recent successors \cite{bopdmd, robust_dmd, optdmd_c} generally utilize variable projection in conjunction with other techniques from optimization in order to compute the spatiotemporal DMD modes. This allows optimized DMD to handle unevenly-sampled snapshots and produce results that are significantly more robust to measurement noise compared to previous approaches like exact DMD \cite{tu_2014}. The variable projection technique extends to generalized objective function formulations for obtaining the components of \eqref{eq:DMDsolution}, enabling  robust DMD \cite{robust_dmd}, which may include model regularizers, constraints, and robust data-fitting penalty functions like the Huber penalty. This flexibility, paired with fast and reliable optimization algorithms have made optimized DMD and its robust extensions highly applicable to real-world applications.

For many applications of DMD, it is often the case that the underlying ground truth system is sparse, both in regards to the number of modes needed to reconstruct the system, as well as the number of spatial collocation points needed to describe each mode. This is largely due to the fact that (1) many systems are inherently low-dimensional despite being observed in high-dimensional state spaces, and (2) many systems consist of spatiotemporal models that are spatially local as opposed to global.  The standard particle in a box  of quantum mechanics, for instance, has both local (discrete mode bound states) and global (continuum) modes which characterize atomic behavior~\cite{griffiths2018}.  Although several sparsity-promoting DMD algorithms have been developed, their goals are not aimed at producing local and global modes. The sparse DMD approach developed by Jovanic et al. \cite{spdmd} promotes sparsity in the amplitudes $b_j$ from the solution \eqref{eq:DMDsolution} with the goal of unveiling an accurate DMD representation with as few spatiotemporal modes as possible. On the other hand, the sparse non-negative DMD approach \cite{sn_dmd} seeks to compute modes $b_j \boldsymbol{\phi}_j$ that are sparse and non-negative in value, with the goal of diminishing the presence of noise and increasing model interpretability for video processing applications. Unfortunately, the algorithm itself utilizes alternating minimizations and second-order proximal Newton methods \cite{prox_newton}, which limits the robustness and flexibility of the method.

As an extension to these sparsity-promoting DMD methods, we introduce the \textit{sparse-mode DMD} approach, which utilizes first-order proximal gradient methods \cite{Parikh2014} in conjunction with variable projection in order to promote sparsity in the DMD modes. In doing so, we effectively combine the robustness and flexibility of the variable projection with the sparsity-promoting formulations of sparse DMD and sparse non-negative DMD. The resulting approach is capable of handling noisy, unevenly-sampled snapshots, and is compatible with a variety of prox-friendly sparse regularizers. Sparse-mode DMD may be thought of as a new instance of the general robust DMD framework~\cite{robust_dmd} that focuses on sparse mode regularizers. A major contribution of this work is the development and deployment of a new fast and accurate sparsity-promoting algorithm that is able to recover mixtures of spatially local as well as spatially global DMD modes from data.

The remainder of this paper will be organized as follows. In Section \ref{sec:background}, we review necessary mathematical background regarding the formulation of DMD and its optimized approaches. In Section \ref{sec:sparse-dmd}, we formulate the sparse-mode DMD algorithm and present a variety of approaches that can be used to implement sparsity promotion. In Section \ref{sec:examples}, we apply our method to a variety of data sets, both synthetic and real-world, while also comparing our results to those of traditional DMD. We additionally explore other interesting aspects of our method in this section, including model accuracy as a function of sparsity strength and regularizer choice, as well as sparsity promotion as a means of de-noising modes and identifying discrete spectra. We then conclude with final remarks and potential method extensions and future research directions in Section \ref{sec:discussion}.

\section{Mathematical Background}
\label{sec:background}

We begin by introducing the mathematical framework and notation necessary to understand optimized DMD and derive the sparse-mode DMD algorithm. First, we introduce the mathematics underlying general DMD. We then introduce variable projection for optimization and review how it can be applied in order to implement the optimized DMD algorithm. We conclude this section by reviewing crucial details regarding the optimized DMD algorithm, which includes the use of Levenberg-Marquardt for eigenvalue updates.

\begin{table}[t]
    \centering
    \begin{tabular}{ll} \toprule
        $x_i$ &\qquad $i$th entry of the vector $\mathbf{x}$. \\ 
        $\mathbf{x}_i$ &\qquad $i$th vector in a collection of vectors $\{\mathbf{x}_k\}_{k\in\mathbb{N}}$. \\ 
        $\mathbf{0}_n$ &\qquad $n$-dimensional vector of zeros. \\ 
        $\mathbf{1}_n$ &\qquad $n$-dimensional vector of ones. \\ 
        $\mathbf{I}_n$ &\qquad $\R^{n \times n}$ identity matrix. \\ 
        $\overline{\mathbf{A}}$ &\qquad Complex conjugate of $\mathbf{A}$. \\
        $\mathbf{A}^\top$ &\qquad Transpose of $\mathbf{A}$. \\
        $\mathbf{A}^*$ &\qquad Complex conjugate transpose of $\mathbf{A}$. \\
        $\mathbf{A}^\dagger$ &\qquad Moore-Penrose pseudoinverse of $\mathbf{A}$. \\ 
        $\mathbf{A}(i, :)$ &\qquad Vector that results from grabbing the $i$th row of $\mathbf{A}$. \\ 
        $\mathbf{A}(:, j)$ &\qquad Vector that results from grabbing the $j$th column of $\mathbf{A}$. \\ 
        $\mathbf{A}(:)$ &\qquad Vector that results from stacking the columns of $\mathbf{A}$ in order. \\ 
        $\mathbf{A} \otimes \mathbf{B}$ &\qquad Kronecker product of $\mathbf{A}$ and $\mathbf{B}$. \\ \toprule
    \end{tabular}
    \caption{Mathematical notation.}
    \label{tab:notation}
\end{table}

\subsection{Dynamic mode decomposition (DMD)}
\label{sec:dmd}

Given data snapshots $\mathbf{x}(t_k) \in \C^n$ collected at times $\{t_k\}_{k=1}^m$ and organized into the data matrix
\begin{equation}
\label{eq:x}
    \mathbf{X} = 
    \begin{bmatrix}
    | & | & & | \\
    \mathbf{x}(t_1) &
    \mathbf{x}(t_2) & 
    \dots & 
    \mathbf{x}(t_m) \\
    | & | & & |
    \end{bmatrix} \in \C^{n \times m},
\end{equation}
the DMD approach generally seeks a rank-$r$ decomposition of the data with the following form:
\begin{equation}
\label{eq:dmd}
    \mathbf{X} \approx 
    \begin{bmatrix}
    | & & | \\
    \boldsymbol{\phi}_1 & \dots & \boldsymbol{\phi}_r \\ 
    | & & |
    \end{bmatrix}
    \begin{bmatrix}
    b_1 \\ & \ddots \\ && b_r
    \end{bmatrix}
    \begin{bmatrix}
    e^{\omega_1 t_1} & \dots & e^{\omega_1 t_m} \\ 
    \vdots & \ddots & \vdots \\
    e^{\omega_r t_1} & \dots & e^{\omega_r t_m}
    \end{bmatrix}
    = \mathbf{\Phi} \text{diag}(\mathbf{b})\Tw.
\end{equation}
In the expression above, the columns $\boldsymbol\phi_i \in \C^n$ denote the spatial modes of the data, the entries $\omega_i \in \C$ give the temporal frequencies associated with the spatial modes, and the entries $b_i \in \R$ give the spatiotemporal mode amplitudes that allow for accurate data reconstructions. By assuming a model of this form, we effectively assume that our system is governed by linear dynamics, where
\begin{equation}
\label{eq:linear}
    \dot{\mathbf{x}}(t) = \mathbf{A} \mathbf{x}(t)
\end{equation}
for $\mathbf{A} \in \R^{n \times n}$. This is because if we assume that our data is governed by \eqref{eq:linear}, we can derive the following closed form expression for our system \cite{leveque, optdmd}, where we use $\mathbf{x}_0 = \mathbf{x}(0)$ to denote the initial condition of our data. Note that we assume that the operator $\mathbf{A}$ is diagonalizable with $\mathbf{A} = \mathbf{\Phi} \mathbf{\Omega} \mathbf{\Phi}^{-1}$ for eigenvectors $\mathbf{\Phi} = \begin{bmatrix} \boldsymbol{\phi}_1, \boldsymbol{\phi}_2, \dots, \boldsymbol{\phi}_n \end{bmatrix}$ and eigenvalues $\mathbf{\Omega} = \text{diag}(\omega_1, \dots, \omega_n)$.
\begin{equation}
    \mathbf{x}(t) = \mathbf{\Phi} e^{\mathbf{\Omega} t} \mathbf{\Phi}^{-1} \mathbf{x}_0
\end{equation}
Since the DMD representation given by \eqref{eq:dmd} for a single snapshot of data is similarly given by $\mathbf{x}(t) = \mathbf{\Phi} e^{\mathbf{\Omega} t} \mathbf{b}$, we immediately see that the spatial modes $\mathbf{\Phi}$ and the frequencies $\w$ respectively correspond with the eigenvectors and the eigenvalues of the operator $\mathbf{A}$ from Equation \eqref{eq:linear}. The amplitudes $\mathbf{b}$ can then be obtained from the relationship $\mathbf{b} = \mathbf{\Phi}^{-1} \mathbf{x}_0$. For this reason, we can think of the DMD problem as an inverse differential equation problem \cite{optdmd}, where given observations of the dynamics $\mathbf{X}$, our goal is to compute the governing linear operator $\mathbf{A}$.

There are a multitude of ways that we can perform DMD in practice. Several methods like exact DMD \cite{tu_2014} and many of its much more robust extensions \cite{tdmd, fbdmd, pidmd, rdmd} implicitly seek the eigendecomposition of the operator $\mathbf{A}$ by using the data matrix $\mathbf{X}$ and a second data matrix $\mathbf{X}'$ whose snapshot columns are advanced one time step $\Delta t$ into the future. Alternatively, we can pose our desired decomposition in \eqref{eq:dmd} as an optimization over the data, such as in \eqref{eq:optdmd}. We may then deploy well-established optimization techniques such as variable projection in order to obtain the spatiotemporal DMD components, as is the approach taken by the optimized DMD algorithm \cite{optdmd} and its extensions \cite{bopdmd, robust_dmd, optdmd_c}. We will explore this approach to DMD in greater depth in the following sections. First, we begin with a general discussion about variable projection, and then we proceed to connect variable projection to DMD and the optimized DMD algorithm.

\subsection{Variable projection}
\label{sec:varpro}

Variable projection \cite{golub2003} as it is described in \cite{sparse_pca} generally deals with the optimization problem
\begin{equation}
\label{eq:min-g}
    \argmin_{\mathbf{A}, \mathbf{B}} g(\mathbf{A}, \mathbf{B})
\end{equation}
for some function $g: \C^{n_1 \times m_1} \times \C^{n_2 \times m_2} \to \R$, $\mathbf{A} \in\C^{n_1 \times m_1}$, $\mathbf{B} \in\C^{n_2 \times m_2}$. We start by rewriting \eqref{eq:min-g} as the following value function optimization problem:
\begin{equation}
\label{eq:min-v}
    \argmin_{\mathbf{B}} \bigg\{ v(\mathbf{B}) := \min_\mathbf{A} g(\mathbf{A}, \mathbf{B}) \bigg\}.
\end{equation}
We introduce \eqref{eq:min-v} because the function $v(\mathbf{B})$ oftentimes possesses an explicit expression, or we can deploy well-established optimization routines in order to obtain the minimizer
\begin{equation}
\label{eq:opt-a}
    \mathbf{A}(\mathbf{B}) = \argmin_{\mathbf{A}} g(\mathbf{A}, \mathbf{B}).
\end{equation}
Using our expression for $\mathbf{A}(\mathbf{B})$, we may then pursue the optimal matrix $\mathbf{B}$ via the optimization
\begin{equation}
\label{eq:opt-b}
    \argmin_{\mathbf{B}} g\left(\mathbf{A}(\mathbf{B}), \mathbf{B}\right),
\end{equation}
which may be solved via any appropriate optimization scheme of the user's choice. If one opts for an iterative scheme like gradient descent or Levenberg-Marquardt, then \eqref{eq:opt-a} may be used to recompute the corresponding optimal $\mathbf{A}(\mathbf{B})$ after each update to $\mathbf{B}$.

\subsection{Variable projection for optimized DMD}
\label{sec:optdmd}

The DMD fitting problem given by \eqref{eq:dmd} can be posed as the following optimization:
\begin{equation}
\label{eq:optdmd}
    \argmin_{\Phib, \: \w}
    \tfrac{1}{2} \| \mathbf{X} - \Phib \Tw \|_F^2,
\end{equation}
where we define the amplitude-scaled mode matrix $\Phib = \mathbf{\Phi} \text{diag}(\mathbf{b})$. We can solve \eqref{eq:optdmd} directly by using variable projection techniques for nonlinear least squares problems \cite{golub1973, golub2003, kaufman1975, oleary2013, golub1979}, as was the approach proposed by Askham and Kutz \cite{optdmd}. In particular, we observe that for a fixed set of DMD eigenvalues $\w$, we can compute the optimal $\Phib$ in \eqref{eq:optdmd} as
\begin{equation}
\label{eq:phi-optdmd}
    \Phibhat = \mathbf{X} \big[ \Tw \big]^\dagger.
\end{equation}
Plugging this expression for $\Phib$ into \eqref{eq:optdmd} then yields the minimization
\begin{equation}
\label{eq:omega-optdmd}
    \argmin_{\w} \tfrac{1}{2} \| \mathbf{X} - \mathbf{X} \big[ \Tw \big]^\dagger \Tw \|_F^2,
\end{equation}
which is a nonlinear least-squares problem that can be solved iteratively via methods like Levenberg-Marquardt \cite{levenberg, marquardt}. This process will be discussed in greater detail in Section \ref{sec:lev-marq}. This method of approaching DMD is referred to as the \textit{optimized DMD} algorithm, and it broadly consists of iteratively stepping towards the optimal $\w$ via \eqref{eq:omega-optdmd} and updating $\Phib$ via \eqref{eq:phi-optdmd} after each update to $\w$. See Algorithm \ref{alg:optdmd} for a full summary of the optimized DMD algorithm.

This approach to DMD has many advantages. In addition to allowing for data snapshots that are unevenly sampled in time, optimized DMD has been shown to optimally suppresses eigenvalue bias introduced by measurement noise. The main disadvantage of this approach is the fact that the optimization given by \eqref{eq:optdmd} is nonlinear and hence prone to convergence failures. Fortunately, this can be remedied by statistical bagging techniques, which stabilize the results of the optimization in addition to producing UQ metrics for $\mathbf{\Phi}, \w, \mathbf{b}$ \cite{bopdmd}. This approach is known as bagging, optimized DMD (BOP-DMD), and is generally one of the most accurate and effective DMD frameworks available to date. The method's amenability to imperfect, high-dimensional measurements has even led to it becoming the backbone of methods like the multi-resolution coherent spatiotemporal scale separation (mrCOSTS) algorithm, which is capable of decomposing complex, real-world data sets that exhibit multi-scale dynamics \cite{mrcosts}. The optimized DMD algorithm has also recently been generalized and extended to encompass control inputs, regularizers, constraints, and penalties beyond the least-squares error norm \cite{optdmd_c, robust_dmd}, all of which are permitted largely due to the algorithm's flexible variable projection-based formulation. As we will see in Section \ref{sec:sparse-dmd}, it is the same variable projection perspective that will allow us to enforce sparsity on the DMD modes.

\begin{algorithm}[t]
\caption{Optimized DMD for \eqref{eq:optdmd}.}
\label{alg:optdmd}
    \begin{algorithmic}
        \Require $\w^0$, $\nu_0$, $k=0$
        \State $\Phib^{0} \gets \mathbf{X} \big[ \mathbf{T}(\w^{0}) \big]^\dagger$
        \While{not converged}
        \State $\boldsymbol{\delta}^k \gets \text{L-M update for} \left\{\argmin_{\w} \tfrac{1}{2} \| \mathbf{X} - \mathbf{X} \big[ \Tw \big]^\dagger \Tw \|_F^2 \right\}$ with damping $\nu_k$ \eqref{eq:omega-optdmd-2} \eqref{eq:lev-marq} \eqref{eq:jacobian-1}
        \If{$F\left(\w^k + \boldsymbol{\delta}^k\right) < F\left(\w^k\right)$ \eqref{eq:omega-optdmd-2}} 
            \State $\w^{k+1} \gets \w^{k} + \boldsymbol{\delta}^{k}$
            \State $\Phib^{k+1} \gets \mathbf{X} \big[ \mathbf{T}(\w^{k+1}) \big]^\dagger$
            \State $\nu_{k+1} \gets \nu_{k} \cdot \max\left\{ \frac{1}{3}, \:1-(2\varrho_k-1)^3 \right\}$ \eqref{eq:nu_update}
            \State $k \gets k + 1$
        \Else
            \State $\nu_{k} \gets \nu_{k} \cdot 2$
        \EndIf
        \EndWhile
        \Ensure $\Phib^k$, $\w^k$
    \end{algorithmic}
\end{algorithm}

\subsection{Levenberg-Marquardt for optimized DMD eigenvalue updates}
\label{sec:lev-marq}

In order to solve the minimization in \eqref{eq:omega-optdmd}, we consider the equivalent optimization
\begin{subequations}
\label{eq:omega-optdmd-2}
\begin{equation}
    \argmin_{\w} \big\{ F(\w) \big\}
     = \argmin_{\w} \left\| \boldsymbol{\rho}(\w) \right\|_2^2
\label{eq:LMobj}
\end{equation}
\begin{equation}
\label{eq:residual}
    \boldsymbol{\rho}(\w) = \mathbf{X}^\top(:) - \left( \mathbf{I}_n \otimes \Tw^\top \right) \left( \mathbf{X} \big[ \Tw \big]^\dagger \right)^\top (:)
\end{equation}
\end{subequations}
with residual $\boldsymbol{\rho}(\w) \in \C^{mn}$. Since \eqref{eq:omega-optdmd-2} is a nonlinear least-squares problem, we deploy an iterative Gauss-Newton-type method in order to step towards a local minimizer of \eqref{eq:omega-optdmd-2} \cite{Madsen2004}:
\begin{equation}
\label{eq:LM-update}
    \w^{k+1} = \w^{k} + \boldsymbol{\delta}^k, \quad k = 0, 1, 2, \dots.
\end{equation}
Note that we use the superscript $k$ to denote iterates of our Gauss-Newton-type algorithm. In this work, we specifically opt for the Levenberg-Marquardt algorithm \cite{levenberg, marquardt} for solving \eqref{eq:omega-optdmd-2}, as this method has shown great success in previous optimized DMD formulations \cite{optdmd, bopdmd}. Traditionally, the $k$th Levenberg-Marquardt update $\boldsymbol{\delta}^k \in \C^r$ is defined to be the minimizer
\begin{equation}
\label{eq:lev-marq-orig}
    \boldsymbol{\delta}^k = \argmin_{\boldsymbol{\delta}} \big\| \boldsymbol{\rho}(\w^k) + \mathbf{J}(\w^k)\boldsymbol{\delta} \big\|_2^2 + \nu_k \|\boldsymbol{\delta}\|_2^2,
\end{equation}
where $\mathbf{J}(\w^k) \in \C^{mn \times r}$ denotes the Jacobian of $\boldsymbol{\rho}$ evaluated at the $k$th eigenvalue iterate $\w^k$, and $\nu_k > 0$ is a damping parameter that is altered based on the current state of the optimization \cite{Madsen2004}. For added consistency with the original optimized DMD derivation \cite{optdmd}, we instead define the $k$th Levenberg-Marquardt update to be the modified minimizer
\begin{align}
\label{eq:lev-marq}
    \boldsymbol{\delta}^k
    &= \argmin_{\boldsymbol{\delta}} \left\| \begin{bmatrix} \boldsymbol{\rho}(\w^k)  \\ 0 \end{bmatrix} + \begin{bmatrix} \mathbf{J}(\w^k) \\ \nu_k \mathbf{M}(\w^k) \end{bmatrix} \boldsymbol{\delta} \right\|_2^2 \\
    &= \argmin_{\boldsymbol{\delta}} \left\| \boldsymbol{\rho}(\w^k) + \mathbf{J}(\w^k)\boldsymbol{\delta} \right\|_2^2 + \nu_k^2 \left\|\mathbf{M}(\w^k)\boldsymbol{\delta} \right\|_2^2,
\end{align}
such that $\mathbf{M}(\w^k) \in \R^{r \times r}$ is a diagonal matrix with entries $\mathbf{M}(\w^k)_{jj} = \|\mathbf{J}(\w^k)(:, j)\|_2$. For our particular residual \eqref{eq:residual}, we can approximate the $j$th column of the Jacobian with
\begin{equation}
\label{eq:jacobian-1}
    \mathbf{J}(:, j)
    := \frac{\partial \boldsymbol{\rho}}{\partial \omega_j}
    \approx -\left((\mathbf{I} - \mathbf{U}\mathbf{U}^*)\left( \mathbf{X} \big[ \Tw \big]^\dagger \frac{\partial \mathbf{T}}{\partial \omega_j} \right)^\top \right)(:),
\end{equation}
where we compute the SVD $\Tw^\top = \mathbf{U}\mathbf{\Sigma}\mathbf{V}^*$ \cite{optdmd}. Hence as long as we are able to generate a reasonable estimate for the eigenvalues $\w^0$, we may utilize the updates defined by \eqref{eq:LM-update} and \eqref{eq:lev-marq} in order to iteratively step towards a local minimizer of \eqref{eq:omega-optdmd-2}. For more details regarding the construction of \eqref{eq:jacobian-1}, and for the full Jacobian expression, we refer readers to the original optimized DMD paper \cite{optdmd}.

Since the damping parameter $\nu_k$ makes a significant impact on the convergence of Levenberg-Marquardt, we take careful note of how we choose this parameter at each step of the algorithm. In particular, large values of $\nu_k$ penalize large updates $\boldsymbol{\delta}$, which is typically ideal during early stages of the algorithm when $\boldsymbol{\omega}^k$ is far from the optimal solution. By contrast, small values of $\nu_k$ allow for large undamped Gauss-Newton updates, which is ideal for fast convergence during later stages of the algorithm \cite{Madsen2004}. In order to update $\nu_k$ accordingly, we deploy the update rule
\begin{equation}
    \nu_{k+1} = \nu_k \cdot \max\left\{ \frac{1}{3}, \:1-(2\varrho_k-1)^3 \right\},
\label{eq:nu_update}
\end{equation}
which is based on the improvement ratio
\begin{equation}
    \varrho_k := \frac{F(\w^k)-F(\w^k+\boldsymbol{\delta}^k)}{L(\mathbf{0}) - L(\boldsymbol{\delta}^k)} = 
    \frac{F(\w^k)-F(\w^k+\boldsymbol{\delta}^k)}{\left(\boldsymbol{\delta}^k\right)^* \left(\nu_k^2 \mathbf{M}(\w^k)^2 \boldsymbol{\delta}^k - \mathbf{J}(\w^k)^*\boldsymbol{\rho}(\w^k)\right)}
\label{eq:ratio}
\end{equation}
as suggested by Madsen at al. \cite{Madsen2004}. Note that $\varrho_k$ specifically describes the ratio between the actual improvement in the objective \eqref{eq:LMobj} and the estimated improvement given by $L(\boldsymbol{\delta}) = \| \boldsymbol{\rho}(\w)+ \mathbf{J}(\w)\boldsymbol{\delta} \|_2^2$, which is based on a linear approximation of $\boldsymbol{\rho}$ about $\w$. In the event that the current damping parameter $\nu_k$ fails to actually improve the objective, we instead increase $\nu_k$ until something works, in lieu of the update rule \eqref{eq:nu_update}. See \cite{Madsen2004} for more details about the Levenberg-Marquardt algorithm and its overall structure.

\subsubsection{Data compression for fast eigenvalue updates}
In the event that the number of features $n$ is very large, the Levenberg-Marquardt update \eqref{eq:lev-marq} can become quite expensive to compute. Hence in practice, we often define some compression matrix $\mathbf{C} \in \C^{p \times n}$ for $p \ll n$ such that the feature-compressed data matrix
\begin{equation}
    \mathbf{\widetilde{X}} := \mathbf{C} \mathbf{X} \in \C^{p \times m}
\end{equation}
is a low-rank approximation of the data that preserves the vast majority of crucial information. In the context of DMD, this is often done using the POD modes of the data \cite{dmd_book}, and is even suggested in the original optimized DMD formulation \cite{optdmd}. Other DMD variants utilize alternative compression techniques, such as matrix sketching \cite{cdmd} or randomized linear algebra \cite{rdmd}. If we define the compressed mode matrix $\mathbf{\widetilde{\Phi}_\mathbf{b}} := \mathbf{C} \Phib$, we can instead solve the compressed optimization
\begin{equation}
    \argmin_{\w}
    \tfrac{1}{2} \| \mathbf{\widetilde{X}} - \mathbf{\widetilde{\Phi}_\mathbf{b}} \Tw \|_F^2,
\end{equation}
or equivalently
\begin{equation}
    \argmin_{\w} \tfrac{1}{2} \| \mathbf{\widetilde{X}} - \mathbf{\widetilde{X}} \big[ \Tw \big]^\dagger \Tw \|_F^2,
\end{equation}
in lieu of \eqref{eq:omega-optdmd} in order to compute the DMD eigenvalues. In practice, doing so significantly reduces computation time with little to no impact on model accuracy.

\subsubsection{Projected Levenberg-Marquardt for constrained eigenvalue updates}
As an alternative to the optimization in \eqref{eq:omega-optdmd}, we can instead consider a constrained variant of the problem, which we define as the following for some $\mathcal{C} \subseteq \C$:
\begin{equation}
\label{eq:omega-constr}
    \argmin_{\w} \tfrac{1}{2} \| \mathbf{X} - \mathbf{X} \big[ \Tw \big]^\dagger \Tw \|_F^2, \quad \text{subject to } \w \in \mathcal{C}.
\end{equation}
There are a variety of reasons why we might want to constrain $\w$. As an example, since the entries of $\w$ define the exponential time evolution of the DMD modes, one might wish to constrain the real components of $\w$ in order to either facilitate oscillatory behavior or deter exponential growth.

One way that we can approximate the solution to the constrained problem \eqref{eq:omega-constr} is by utilizing what is known as projected Levenberg-Marquardt \cite{kanzow2004}, which uses the updates
\begin{equation}
\label{eq:LM-update-constr}
    \w^{k+1}_\mathcal{C} = \mbox{proj}_{\mathcal{C}} ( \w^{k}_\mathcal{C} - \boldsymbol{\delta}^k )
\end{equation}
as opposed to the updates given by \eqref{eq:LM-update}. Note that $\mbox{proj}_{\mathcal{C}}(\mathbf{z})$ denotes the projection of $\mathbf{z}$ onto the region $\mathcal{C}$, and that we use the subscript $\mathcal{C}$ to denote results of the constrained optimization. Also notice that the projected updates still utilize the steps defined by \eqref{eq:lev-marq}. Projected Levenberg-Marquardt is simple to implement, computationally efficient, and quite effective in practice for a variety of feasible regions $\mathcal{C}$, as we will demonstrate in Section \ref{sec:examples}. However, it is crucial to note that projected Levenberg-Marquardt is not guaranteed to converge globally \cite{kanzow2004}. Hence in the event that the projected updates \eqref{eq:LM-update-constr} yield worse results than the unprojected updates \eqref{eq:LM-update}, one must opt for alternative methods for enforcing the constraints described in \eqref{eq:omega-constr}. For example, one might deploy proximal gradient methods as described in the robust DMD formulation \cite{robust_dmd}, or one might deploy pre-built constrained linear least-squares solvers \cite{optdmd_c}.

\section{Optimized DMD with Sparse Modes}
\label{sec:sparse-dmd}

In this work, we are specifically interested in the following reformulation of the DMD optimization:
\begin{equation}
\label{eq:optdmd_sparse}
    \argmin_{\Phib, \: \w}
    \tfrac{1}{2} \| \mathbf{X} - \Phib \Tw \|_F^2 + \psi(\Phib),
\end{equation}
where the function $\psi$ denotes a generic sparse regularizer. In doing so, we seek a decomposition of our data with the form given by \eqref{eq:dmd} while also promoting sparsity in the spatial modes. This reformulation requires that we modify the alternating updates discussed in Section \ref{sec:optdmd}. In this section, we highlight several different approaches that can be taken in order to solve \eqref{eq:optdmd_sparse}. In particular, we discuss and derive implementations involving (1) proximal gradient methods, (2) sequential thresholding methods, and (3) global-local mode separation.

\begin{algorithm}[t]
\caption{Sparse-mode DMD for \eqref{eq:optdmd_sparse}.}
\label{alg:sparse-mode-dmd}
    \begin{algorithmic}
        \Require $\psi$, $\w^0$, $\nu_0$, $k=0$
        \State $\Phib^{0} \gets \text{Sparse-mode-update}\left(\mathbf{X} \big[ \mathbf{T}(\w^{0}) \big]^\dagger, \w^0, \psi \right)$
        \While{not converged}
        \If{\text{using de-biased mode updates}}
            \State $\boldsymbol{\delta}^k \gets \text{L-M update for} \left\{\argmin_{\w} \tfrac{1}{2} \| \mathbf{X} - \mathbf{X} \big[ \Tw \big]^\dagger \Tw \|_F^2 \right\}$with damping $\nu_k$ \eqref{eq:lev-marq}\eqref{eq:residual-2}\eqref{eq:jacobian-3}
        \Else
            \State $\boldsymbol{\delta}^k \gets \text{L-M update for} \left\{\argmin_{\w} \tfrac{1}{2} \| \mathbf{X} - \Phib^{k} \Tw \|_F^2 \right\}$with damping $\nu_k$ \eqref{eq:lev-marq}\eqref{eq:residual-2}\eqref{eq:jacobian-2}
        \EndIf
        \State $\Phibhat \gets \text{Sparse-mode-update}\left(\Phib^{k}, \w^{k}+\boldsymbol{\delta}^k, \psi \right)$
        \If{$F_s\left(\w^k + \boldsymbol{\delta}^k; \Phibhat\right) < F_s\left(\w^k; \Phib^k\right)$ \eqref{eq:LMobj-2}} 
            \State $\w^{k+1} \gets \w^{k} + \boldsymbol{\delta}^{k}$
            \State $\Phib^{k+1} \gets \Phibhat$
            \State $\nu_{k+1} \gets \nu_{k} \cdot \frac{1}{3}$
            \State $k \gets k + 1$
        \Else
            \State $\nu_{k} \gets \nu_{k} \cdot 2$
        \EndIf
        \EndWhile
        \Ensure $\Phib^k$, $\w^k$
    \end{algorithmic}
\end{algorithm}

\subsection{Proximal gradient methods for sparse mode updates}
\label{sec:sparse-dmd-prox}

For a fixed set of eigenvalues $\w$, the optimal modes $\Phibhat$ must now solve the following optimization:
\begin{equation}
\label{eq:phi_sparse}
    \Phibhat = \argmin_{\Phib}
    \tfrac{1}{2} \| \mathbf{X} - \Phib \Tw \|_F^2 + \psi(\Phib).
\end{equation}
Assuming that the rows of the matrix $\Phib$ are independent, as was the approach utilized in \cite{robust_dmd}, we can split the optimization in \eqref{eq:phi_sparse} amongst the rows $\Phib(i, :) \in \C^r$. In doing so, we obtain following set of decoupled optimization problems, all of which can be solved in parallel:
\begin{equation}
\label{eq:phi_sparse_cols}
    \Phibhat(i, :) = \argmin_{\mathbf{z}}
    \tfrac{1}{2} \| \mathbf{X}(i, :) - \Tw^\top \mathbf{z} \|_2^2 + \psi(\mathbf{z}), \quad i = 1, 2, \dots, n.
\end{equation}
Notice that we can succinctly express the optimization problem in \eqref{eq:phi_sparse_cols} as the following:
\begin{equation}
\label{eq:Az=b}
    \Phibhat(i, :) = \argmin_{\mathbf{z}}
    f_i(\mathbf{z}) + \psi(\mathbf{z}),
\end{equation}
\begin{equation*}
    f_i(\mathbf{z}) := \tfrac{1}{2} \| \mathbf{A} \mathbf{z} - \mathbf{b}_i\|_2^2, \quad \mathbf{A} := \Tw^\top \in \C^{m \times r}, \quad \mathbf{b}_i := \mathbf{X}(i, :) \in \C^m.
\end{equation*}
The function $f_i: \C^r \to \R$ denotes the smooth part of our objective, while our regularizer $\psi$ is potentially non-smooth. Notice that $f_i$ is in fact $\beta$-smooth for fixed $\w$, as the complex Wirtinger gradient of $f_i$ is given by the following, as derived in previous works \cite{cr_calc, complex_grad_1, complex_grad_2}.
\begin{equation}
\label{eq:grad}
    \nabla f_i(\mathbf{z})
    = \mathbf{A}^* (\mathbf{A} \mathbf{z} -\mathbf{b}_i)
\end{equation}
Note that we generally define the complex Wirtinger gradient via
\begin{align}
    \frac{\partial f}{\partial \mathbf{z}} &= \frac{1}{2} \left( \frac{\partial f}{\partial \mathbf{x}} - i \frac{\partial f}{\partial \mathbf{y}} \right) & 
    \frac{\partial f}{\partial \overline{\mathbf{z}}} &= \frac{1}{2} \left( \frac{\partial f}{\partial \mathbf{x}} + i \frac{\partial f}{\partial \mathbf{y}} \right)
\end{align}
for functions $f(\mathbf{z}, \overline{\mathbf{z}})$ that are complex differentiable with respect to both $\mathbf{z} \in \C^r$ and the complex conjugate $\overline{\mathbf{z}}$, where $\mathbf{z} = \mathbf{x} + i \mathbf{y}, \mathbf{x}, \mathbf{y} \in \R^{r}$. From \eqref{eq:grad}, we can establish the following upper bound for any $\mathbf{z}_1, \mathbf{z}_2 \in \C^{r}$ and determine that $f_i$ is $\beta$-smooth for $\beta = \| \mathbf{A} \|_2^2$ by definition. This in turn provides us with a consistent and reliable step size when applying proximal gradient methods \cite{Parikh2014} to solve \eqref{eq:phi_sparse_cols}.
\begin{align}
\label{eq:beta}
    \left\| \nabla f_i(\mathbf{z}_1) - \nabla f_i(\mathbf{z}_2) \right\| 
    &= \| \mathbf{A}^* \mathbf{A} (\mathbf{z}_1 - 
    \mathbf{z}_2) \| \nonumber \\ 
    &\leq \| \mathbf{A}^* \mathbf{A} \|_2 \| \mathbf{z}_1 - \mathbf{z}_2 \|
    = \| \mathbf{A} \|_2^2 \| \mathbf{z}_1 - \mathbf{z}_2 \|
\end{align}
Hence if we choose a sparse regularizer $\psi$ such that $\psi$ is convex, we can deploy proximal gradient methods in order to solve the optimization problem in \eqref{eq:phi_sparse}, and even if we do not choose a convex regularizer, we can still generally deploy a proximal gradient-based pipeline as long as we can compute the proximal operator associated with our regularizer.

For example, one might deploy the proximal gradient algorithm, whose parallelized iterative updates for solving \eqref{eq:phi_sparse_cols} are defined as follows for the step size $\gamma = 1 / \beta$. This then yields the simple pipeline outlined in Algorithm \ref{alg:prox-grad}. Notice that we use the gradient \eqref{eq:grad} to define our updates, while we use the superscript $\ell$ to denote iterations of the prox-gradient algorithm. We also use $\mbox{prox}_{\gamma \psi}(\cdot)$ to denote the proximal operator associated with $\psi$. For more information on proximal operators and various sparse regularizers, please see Section \ref{sec:prox}.
\begin{align}
\label{eq:prox-grad}
    \left( \Phib^{\ell+1} \right)^\top
    &= \mbox{prox}_{\gamma \psi} \left[ \left( \Phib^{\ell} \right)^\top - \gamma \mathbf{A}^* \left( \mathbf{A} \left( \Phib^{\ell} \right)^\top - \mathbf{X}^\top \right) \right]
\end{align}
Alternatively, one might opt for a more sophisticated proximal method for updating the mode matrix $\Phib$. For example, one might use an accelerated first-order method like FISTA \cite{fista}, that or one might opt for a higher-order proximal Newton-type method \cite{prox_newton}, as was utilized for the sparse-nonnegative DMD (SN-DMD) approach \cite{sn_dmd}. It is worth noting however that first-order proximal methods tend to utilize easy-to-compute, closed-form proximal operators, which consequently allows us to efficiently utilize a wide variety of sparse regularizers. We hence utilize FISTA in our own implementation of sparse-mode DMD.

Once we optimize for the matrix $\Phib$, we may then proceed to optimize over the eigenvalues $\w$ for fixed $\Phib$ via the following optimization. Notice that the following objective is smooth, as the non-smooth regularizer term $\psi$ drops out of the optimization in \eqref{eq:optdmd_sparse} once we fix the matrix $\Phib$.
\begin{equation}
\label{eq:omega}
    \argmin_{\w}
    \tfrac{1}{2} \| \mathbf{X} - \Phibhat \Tw \|_F^2
\end{equation}
Unfortunately, since the optimal matrix $\Phibhat$ no longer has the closed-form representation given by $\Phibhat = \mathbf{X} [\Tw]^\dagger$, we can no longer utilize the Levenberg-Marquardt step formulated for the original optimized DMD algorithm given by \eqref{eq:jacobian-1}. We can however still utilize the Levenberg-Marquardt algorithm if we allow the optimal matrix $\Phibhat$ to take on a more general structure. More specifically, if we instead assume that the Levenberg-Marquardt residual $\boldsymbol{\rho}(\w)$ is now given by
\begin{equation}
\label{eq:residual-2}
    \boldsymbol{\rho}\left(\w; \: \Phibhat\right) = \mathbf{X}^\top(:) - \left( \mathbf{I}_n \otimes \Tw^\top \right) \left( \Phibhat \right)^\top (:)
\end{equation}
instead of \eqref{eq:residual}, the $j$th column of the Jacobian of our residual now becomes the following, with the remainder of the Levenberg-Marquardt steps from Section \ref{sec:lev-marq} remaining largely the same.
\begin{equation}
\label{eq:jacobian-2}
    \mathbf{J}(:, j)
    := \frac{\partial \boldsymbol{\rho}}{\partial \omega_j}
    = -\left( \Phibhat \frac{\partial \mathbf{T}}{\partial \omega_j} \right)^\top(:).
\end{equation}

We do note however that the introduction of sparsity slightly impacts how we update the damping parameter $\nu_k$. In particular, since our residual $\boldsymbol{\rho}$ is parameterized by the mode matrix $\Phibhat$ in the sparse-mode problem, we may adopt the perspective that our Levenberg-Marquardt optimization is actually
\begin{equation}
    \argmin_{\w} \left\{ F_s(\w; \Phibhat) \right\}
     = \argmin_{\w} \left\| \boldsymbol{\rho}(\w; \Phibhat) \right\|_2^2 + \psi(\Phibhat).
\label{eq:LMobj-2}
\end{equation}
This raises a few complications, as the matrix $\Phibhat$ is constantly being updated across iterations and our improvement is no longer neatly summarized by $\varrho_k$ \eqref{eq:ratio}. Nonetheless, we may still adopt a reasonable damping parameter updating scheme, simply by deploying the same line of reasoning used to update $\nu_k$ in the un-regularized case: decrease $\nu_k$ if stepping in the direction $\boldsymbol{\delta}^k$ improves our objective, and increase $\nu_k$ if we fail to improve the objective. More specifically, we decrease $\nu_k$ by a factor of $3$ if updating $\w^k$ by $\boldsymbol{\delta}^k$ improves $F_s$, or we increase $\nu_k$ by a factor of $2$ and recompute $\boldsymbol{\delta}^k$ until we actually improve the objective $F_s$. Note that this approach is similar to the updating strategy originally proposed by Marquardt \cite{marquardt, Madsen2004}. We hence utilize \eqref{eq:residual-2}, \eqref{eq:jacobian-2}, and \eqref{eq:LMobj-2} along with the Levenberg-Marquardt steps outlined in Section \ref{sec:lev-marq} in order to update $\w$ when performing sparse-mode DMD. See Algorithms \ref{alg:sparse-mode-dmd} and \ref{alg:prox-grad} for a full method outline.

\begin{algorithm}[t]
\caption{Proximal gradient for sparse mode updates.}
\label{alg:prox-grad}
    \begin{algorithmic}
        \Require $\Phib^{0}$, $\w$, $\psi$, $\ell=0$
        \State $\mathbf{A} \gets \mathbf{T}(\w)^\top$
        \State $\gamma \gets 1 / \| \mathbf{A} \|_2^2$
        \While{$\Phib^{\ell}$ not converged}
        \State $\left( \Phib^{\ell+1} \right)^\top
            \gets \mbox{prox}_{\gamma \psi} \left[ \left( \Phib^{\ell} \right)^\top - \gamma \mathbf{A}^* \left( \mathbf{A} \left( \Phib^{\ell} \right)^\top - \mathbf{X}^\top \right) \right]$
        \State $\ell \gets \ell + 1$
        \EndWhile
        \Ensure $\Phib^\ell$
    \end{algorithmic}
\end{algorithm}

\subsection{Sparse relaxed regularized regression (SR3) for fast, debiased updates}
\label{sec:sparse-dmd-sr3}

As an alternative to directly solving \eqref{eq:optdmd_sparse}, we can instead consider the relaxed optimization
\begin{equation}
\label{eq:optdmd_sparse_reg}
    \argmin_{\Phib, \: \w, \: \mathbf{W}}
    \tfrac{1}{2} \| \mathbf{X} - \Phib \Tw \|_F^2 + \tfrac{1}{2\eta} \| \Phib - \mathbf{W} \|_F^2 + \psi(\mathbf{W}),
\end{equation}
where we introduce the auxiliary matrix $\mathbf{W} \in \C^{n \times r}$ upon which we apply sparsity. We additionally introduce the parameter $\eta > 0$, which controls the gap between our amplitude-scaled mode matrix $\Phib$ and the auxiliary matrix $\mathbf{W}$ such that the relaxed optimization \eqref{eq:optdmd_sparse_reg} closely approximates the original optimization \eqref{eq:optdmd_sparse} for sufficiently small $\eta$. This approach is known as sparse relaxed regularized regression (SR3), and has been shown to improve runtime, robustness, and conditioning for a variety of sparse regression applications \cite{sr3, sr3_sindy}.

In order to solve for the optimal modes of the relaxed optimization \eqref{eq:optdmd_sparse_reg}, we must now solve for both the optimal mode matrix $\Phibhat$ as well as the optimal auxiliary matrix $\mathbf{\hat{W}}$ via
\begin{equation}
\label{eq:phi_w_sparse_reg}
    \Phibhat, \mathbf{\hat{W}} = \argmin_{\Phib, \: \mathbf{W}}
    \tfrac{1}{2} \| \mathbf{X} - \Phib \Tw \|_F^2 + \tfrac{1}{2\eta} \| \Phib - \mathbf{W} \|_F^2 + \psi(\mathbf{W}).
\end{equation}
For a fixed $\w$ and $\mathbf{W}$, our optimal mode matrix $\Phibhat$ is now given by
\begin{equation}
\label{eq:phi_sparse_reg}
    \Phibhat = \argmin_{\Phib}
    \tfrac{1}{2} \| \mathbf{X} - \Phib \Tw \|_F^2 + \tfrac{1}{2\eta} \| \Phib - \mathbf{W} \|_F^2.
\end{equation}
If we again assume that the rows of the matrix $\Phib$ are independent, as was done in Section \ref{sec:sparse-dmd-prox}, we obtain the following set of optimization problems for $\mathbf{w}_i = \mathbf{W}(i, :)$ and $\mathbf{A}$, $\mathbf{b}_i$ defined in \eqref{eq:Az=b}.
\begin{align}
\label{eq:Az=b_smooth}
    \Phibhat(i, :)
    &= \argmin_{\mathbf{z}} \tfrac{1}{2} \| \mathbf{A} \mathbf{z} - \mathbf{b}_i \|_2^2 + \tfrac{1}{2\eta} \| \mathbf{z} - \mathbf{w}_i \|_2^2 \\ 
    &= \argmin_{\mathbf{z}} g_i(\mathbf{z}), \quad i = 1, 2, \dots, n.
\end{align}
Since this objective is smooth and differentiable with respect to $\mathbf{z}$, we can now solve \eqref{eq:Az=b_smooth} via least-squares. In particular, we note that the gradient of the objective in \eqref{eq:Az=b_smooth} is now given by
\begin{equation}
\label{eq:sr3_grad}
    \nabla g_i(\mathbf{z}) = \mathbf{A}^* (\mathbf{A} \mathbf{z} -\mathbf{b}_i) + \tfrac{1}{\eta}(\mathbf{z} -\mathbf{w}_i),
\end{equation}
in which case setting \eqref{eq:sr3_grad} equal to the zero vector and solving for $\mathbf{z}$ yields the following expression for the optimal vector $\hat{\mathbf{z}}$. In other words, the rows of the optimal mode matrix $\Phibhat$ are given by:
\begin{equation}
\label{eq:sr3_phi}
    \Phibhat(i, :) = \big( \mathbf{A}^* \mathbf{A} + \tfrac{1}{\eta} \mathbf{I}_r \big)^\dagger \big( \mathbf{A}^* \mathbf{b}_i + \tfrac{1}{\eta} \mathbf{w}_i \big).
\end{equation}
Conversely, notice that for a fixed $\w$ and $\Phib$, the rows of the optimal auxiliary matrix $\mathbf{\hat{W}}$ must solve the following optimization if we assume row independence, which we note is precisely the definition of the proximal operator. See Section \ref{sec:prox} for more details.
\begin{align}
    \mathbf{\hat{W}}(i, :)
    \label{eq:phi_sparse_sr3}
    &= \argmin_{\mathbf{w}} \tfrac{1}{2\eta} \| \Phib(i, :) - \mathbf{w} \|_2^2 + \psi(\mathbf{w}) \\ 
    \label{eq:sr3_w}
    &= \mbox{prox}_{\eta \psi} \big( \Phib(i, :) \big)
\end{align}
We may hence utilize the alternating updates given by \eqref{eq:sr3_phi} and \eqref{eq:sr3_w} in order to update the mode matrix $\Phib$ instead of the proximal gradient updates given in Section \ref{sec:sparse-dmd-prox}. This approach often vastly improves runtime in practice and is similar to performing sequentially thresholded least squares (STLSQ) \cite{sr3_sindy, sindy}. Furthermore, since $\mathbf{\hat{W}}$ reveals the optimal locations of the non-zero entries of $\Phib$, we can use our resulting $\mathbf{\hat{W}}$ from \eqref{eq:phi_w_sparse_reg} to perform a final debiasing step in which we re-compute the entries of $\Phibhat$ via a least squares fit, but performed only on the entries given by $\mathbf{\hat{W}}$. This in turn prompts us to utilize Levenberg-Marquardt updates with the Jacobian approximation
\begin{equation}
\label{eq:jacobian-3}
    \mathbf{J}(:, j)
    := \frac{\partial \boldsymbol{\rho}}{\partial \omega_j}
    \approx -\left((\mathbf{I} - \mathbf{U}\mathbf{U}^*)\left( \Phibhat \frac{\partial \mathbf{T}}{\partial \omega_j} \right)^\top \right)(:),
\end{equation}
which is more similar to the approximation used in the original optimized DMD algorithm. This process overall gives rise to Algorithms \ref{alg:sparse-mode-dmd} and \ref{alg:sr3}.

\begin{algorithm}[t]
\caption{SR3 for sparse mode updates.}
\label{alg:sr3}
    \begin{algorithmic}
        \Require $\Phib^0$, $\w$, $\psi$, $\eta$, $\ell=0$
        \State $\mathbf{W}^{0} \gets \Phib^0$
        \State $\mathbf{A} \gets \mathbf{T}(\w)^\top$
        \While{$\mathbf{W}^{\ell}$ not converged}
        \State $\left( \Phib^{\ell+1} \right)^\top
            \gets \big( \mathbf{A}^* \mathbf{A} + \tfrac{1}{\eta} \mathbf{I}_r \big)^\dagger \big( \mathbf{A}^* \mathbf{X}^\top + \tfrac{1}{\eta} (\mathbf{W}^{\ell})^\top \big)$
        \State $\left( \mathbf{W}^{\ell+1} \right)^\top
            \gets \mbox{prox}_{\eta\psi} \left[ \left( \Phib^{\ell+1} \right)^\top \right]$
        \State $\ell \gets \ell + 1$
        \EndWhile
        \For{$i = 1, 2, \dots n$}
            \State $\mathbf{\Phi}_\mathbf{b}^{\ell}(i, :) \gets \mathbf{0}_r$
            \State $J_{\text{active}} \gets j \in \{1, 2, \dots, r\}$ such that $\mathbf{W}^{\ell}(i, j) \neq 0$
            \State $\mathbf{\Phi}_\mathbf{b}^{\ell}(i, J_{\text{active}}) \gets \mathbf{X}(i, :) \big[ \mathbf{T}(\boldsymbol{\omega})(J_{\text{active}}, :) \big]^\dagger$
        \EndFor
        \Ensure $\Phib^\ell$
    \end{algorithmic}
\end{algorithm}

\subsection{Sparse-mode DMD with global and local separation}
\label{sec:sparse-dmd-gl}

Although the optimization discussed in Section \ref{sec:sparse-dmd} is quite effective for many data sets, we note that for several real-world data sets, there often exists at least one spatiotemporal mode of the system that is truly spatially global. This could be the background mode of a surveillance video, seasonal fluctuations in climate data, or simply any spatiotemporal feature that dominates the dynamics at a global scale. Hence for such datasets, it is more reasonable to regularize not the entire mode matrix $\Phib$, but rather a subset of $\Phib$ that accounts for the spatially local modes of the data:
\begin{equation}
\label{eq:optdmd_sparse_gl}
    \argmin_{\Phib, \: \w}
    \tfrac{1}{2} \| \mathbf{X} - \Phib \Tw \|_F^2 + \psi(\mathbf{\Phi}_{\text{local}}).
\end{equation}
Here we define the following column-wise partition of the amplitude-scaled mode matrix:
\begin{equation}
    \Phib = 
    \begin{bmatrix}
        \mathbf{\Phi}_{\text{global}} & \mathbf{\Phi}_{\text{local}}
    \end{bmatrix} \in \C^{n \times r},
\end{equation}
where the matrices $\mathbf{\Phi}_{\text{global}} \in \C^{n \times r_g}$ and $\mathbf{\Phi}_{\text{local}} \in \C^{n \times r_\ell}$ account for the spatially global modes and the spatially local modes respectively. Defining the same global-local partition for the vectors defined in \eqref{eq:phi_sparse_cols} and \eqref{eq:phi_sparse_sr3}, notice that the introduction of a global-local split leads to the following alternative sparse optimizations for our prox-gradient and SR3 methods respectively:
\begin{align}
    \Phibhat(i, :) &= \argmin_{\mathbf{z}}
    \tfrac{1}{2} \| \mathbf{A} \mathbf{z} - \mathbf{b}_i \|_2^2 + \psi(\mathbf{z}_{\text{local}}) \\ 
    \mathbf{\hat{W}}(i, :)
    &= \argmin_{\mathbf{w}} \tfrac{1}{2\eta} \| \Phib(i, :) - \mathbf{w} \|_2^2 + \psi(\mathbf{w}_{\text{local}}), \quad i = 1, 2, \dots, n.
\end{align}
Assuming that our sparse regularizer $\psi$ is separable, that is $\psi$ can be written as the sum
\begin{equation}
\label{eq:sep}
    \psi(\mathbf{z}) = \sum_{j=1}^{r} \psi(z_j)
\end{equation}
and its proximal operator may be written as
\begin{equation}
    \left( \mbox{prox}_{\psi}(\mathbf{z}) \right)_j
    = \mbox{prox}_{\psi}(z_j),
\end{equation}
as is the case for many common sparse regularizers such as the $\ell_0$ norm and the $\ell_1$ norm \cite{Parikh2014}, we can alternatively express our local regularization as a new regularization function $\psi_{\text{GL}}$, which we define to be the following weighted sum of scalar functions:
\begin{equation}
    \psi_{\text{GL}}(\mathbf{z})
    := \psi(\mathbf{z}_{\text{local}})
    = \sum_{j = 1}^{r} \mathbbm{1} \{ \text{mode $j$ is local} \} \, \psi(z_j)
    = \sum_{j = 1}^{r} \psi_j(z_j).
\end{equation}
Since $\psi_{\text{GL}}$ is itself a separable function, we find that the $j$th element of the proximal operator of $\psi_{\text{GL}}$ may be expressed as the following, where we utilize the fact that the proximal operator of the zero function $f(x) = 0$ simply returns its input.
\begin{equation}
    \left( \mbox{prox}_{\psi_{\text{GL}}}(\mathbf{z}) \right)_j
    = \mbox{prox}_{ \psi_j}(z_j)
    = \begin{cases}
        \mbox{prox}_{\psi}(z_j) & \text{mode $j$ is local} \\ 
        z_j & \text{mode $j$ is global}
    \end{cases}
\end{equation}
In other words, we find that for separable regularizers of the form \eqref{eq:sep}, the proximal operator associated with applying a global-local split \eqref{eq:optdmd_sparse_gl} simply involves applying the same proximal operator as before, but only to the modes that are considered local, hence the only modification that needs to be made to our algorithms is to utilize $\mbox{prox}_{\psi_{\text{GL}}}(\cdot)$ instead of $\mbox{prox}_{\psi}(\cdot)$.

One way that we can determine if a mode is global as opposed to local is by examining the columns of $\Phib^{0}$ and manually selecting modes to exclude from regularization. This tends to work well for systems with few spatiotemporal modes, however in the event that one would like to apply DMD to a very high-dimensional system that consists of many spatiotemporal modes, this process becomes quite tedious and unreliable. Hence in this work, we deploy the following condition using the parameters $\tau_{\text{active}}$ and $\tau_{\text{global}}$ in order to determine whether or not a mode is global at the $k$th algorithm iteration:
\begin{equation}
    \begin{cases}
        \text{mode $j$ is global} & \sum_{i=1}^{n} \mathbbm{1} \left\{ | \Phib^{k}(i, j) | > \tau_{\text{active}} \cdot \| \Phib^{k} \|_{\text{max}} \right\} > \tau_{\text{global}} \cdot n, \\ 
        \text{mode $j$ is local} & \text{else.}
    \end{cases}
\end{equation}
Notice that we use the max norm of the entire matrix $\Phib^{k}$ as opposed to the column-wise infinity norm in order to avoid erroneously excluding low-amplitude, noise-dominated modes from sparsification. For all examples in Section \ref{sec:examples}, we use $\tau_{\text{active}} = 0.1$ and $\tau_{\text{global}} = 0.5$ unless otherwise specified, largely due to these parameters working well in practice for our particular test data sets. We note however that these parameters may be determined via more sophisticated techniques like cross-validation, and should generally be tailored to the particular data set that one is examining.

\subsection{Sparse regularizers and their proximal operators}
\label{sec:prox}

\bgroup
\begin{table}[t]
    \centering
    \begin{tabular}{ll} \toprule
        Regularizer $\psi$ & Proximal Operator \\ \toprule \\[-1em]
        $\| \mathbf{x} \|_0 := \sum_{i=1}^n \mathbbm{1} \{ x_i \neq 0 \}$ &
        $\left(\mbox{prox}_{\lambda \|\cdot \|_0}(\mathbf{y})\right)_i = 
        \begin{cases}
            y_i, & |y_i|^2 > 2 \lambda \\
            0, & \text{else}
        \end{cases}$ \\ \\[-1em]
        $\| \mathbf{x} \|_1 := \sum_{i=1}^n |x_i|$ &
        $\left(\mbox{prox}_{\lambda \|\cdot \|_1}(\mathbf{y})\right)_i = 
        \begin{cases}
            \mbox{sgn}(y_i)(|y_i| - \lambda), & |y_i| > \lambda \\ 
            0, & \text{else}
        \end{cases}$ \\ \\[-1em]
        $\| \mathbf{x} \|_0 + \| \mathbf{x} \|_2^2$ &
        $\left(\mbox{prox}_{\lambda_1 \|\cdot \|_0 + \lambda_2 \|\cdot \|_2^2}(\mathbf{y})\right)_i = 
        \begin{cases}
            y_i / (1 + 2\lambda_2), & |y_i|^2 > 2\lambda_1 (1 + 2 \lambda_2) \\
            0, & \text{else}
        \end{cases}$ \\ \\[-1em]
        $\| \mathbf{x} \|_1 + \| \mathbf{x} \|_2^2$ & 
        $\left(\mbox{prox}_{\lambda_1 \|\cdot \|_1 + \lambda_2 \|\cdot \|_2^2}(\mathbf{y})\right)_i = 
        \begin{cases}
            \mbox{sgn}(y_i)(|y_i| - \lambda_1)/(1 + 2\lambda_2), & |y_i| > \lambda_1 \\ 
            0, & \text{else}
        \end{cases}$ \\ \\[-1em] \toprule
    \end{tabular}
    \caption{All sparse regularizers implemented and considered in this work, along with their corresponding proximal operators.}
    \label{tab:regularizers}
\end{table}

We define the proximal operator associated with $\psi: \mathbb{C}^n \rightarrow\overline {\mathbb{R}}$ as follows for $\lambda > 0$ and $\mathbf{x}, \mathbf{y} \in \C^n$.
\begin{equation}
\label{eq:prox-def}
    \mbox{prox}_{\lambda \psi}(\mathbf{y}) := \argmin_{\mathbf{x}} \tfrac{1}{2\lambda}\|\mathbf{x}-\mathbf{y}\|_2^2 + \psi(\mathbf{x})
\end{equation}
By solving \eqref{eq:prox-def} for a regularizer $\psi$, we can define the corresponding proximal operator $\mbox{prox}_{\lambda\psi}(\cdot)$, which oftentimes possesses a closed form and this easy and efficient to compute. Here, we discuss the regularizers explored in this work, along with their proximal operators and their unique sparsity-promoting properties. See Table \ref{tab:regularizers} for a list of all of the sparse regularizers that are implemented as a part of this work.

The primary regularizer utilized throughout this work is the $\ell_1$ norm, which is defined for vectors $\mathbf{x} \in \C^n$ as the sum of the magnitudes of all of the elements contained within $\mathbf{x}$. The $\ell_1$ norm is a common choice for many sparsity-promoting algorithms thanks to its convexity and its ability to approximate the $\ell_0$ norm. The proximal operator associated with this norm is often referred to as \textit{soft-thresholding}. See Table \ref{tab:regularizers} for a definition. We also consider the use of the $\ell_0$ norm regularizer in this work, which is defined for vectors $\mathbf{x} \in \C^n$ as the number of nonzero elements within the vector. Unlike the $\ell_1$ norm, the $\ell_0$ norm is \textit{not} convex, hence its use as a regularizer has the potential to lead to local, sub-optimal solutions. Regardless, the $\ell_0$ norm possesses several advantages, as it is able to directly penalize dense solutions since it computes the number of non-zero elements as opposed to the magnitudes of elements. Its proximal operator \textit{hard-thresholding} is also very easy to compute, which makes it quite convenient and suitable for most proximal gradient-based methods.

Although not primarily used and demonstrated in this work, we still implement and consider a few other sparse regularizers. Of these is $\ell_2$ norm regularization and the elastic net regularizer \cite{elastic_net}, which combines the $\ell_1$ norm penalty with the quadratic penalty term $\| \mathbf{x} \|_2^2$ and whose proximal operator is given by scaled soft-thresholding \cite{Parikh2014}. Likewise, we consider the variant of this penalty that uses the $\ell_0$ norm in place of the $\ell_1$ norm, and whose proximal operator is similarly given by scaled hard-thresholding \cite{sparse_pca}.

\section{Applications and Examples}
\label{sec:examples}

In this section, we apply sparse-mode DMD to a variety of example data sets. We begin by analyzing a simple synthetic data set that consists of one spatially global mode and two spatially local modes. In doing so, we demonstrate the general impact of the application of sparsity in the DMD pipeline. We additionally use this example to highlight how different algorithms and sparse regularizers impact the results of sparse-mode DMD. Next we apply sparse-mode DMD to optical waveguide data induced by a finite potential well. This example specifically demonstrates the results that are derived when attempting to model systems that possess both discrete and continuous spectra. As a concluding example, we apply sparse-mode DMD to real-world sea surface temperature data.

\subsection{Synthetic video data}
\label{sec:ex1}

\begin{figure}
    \centering
    \includegraphics[width=\textwidth]{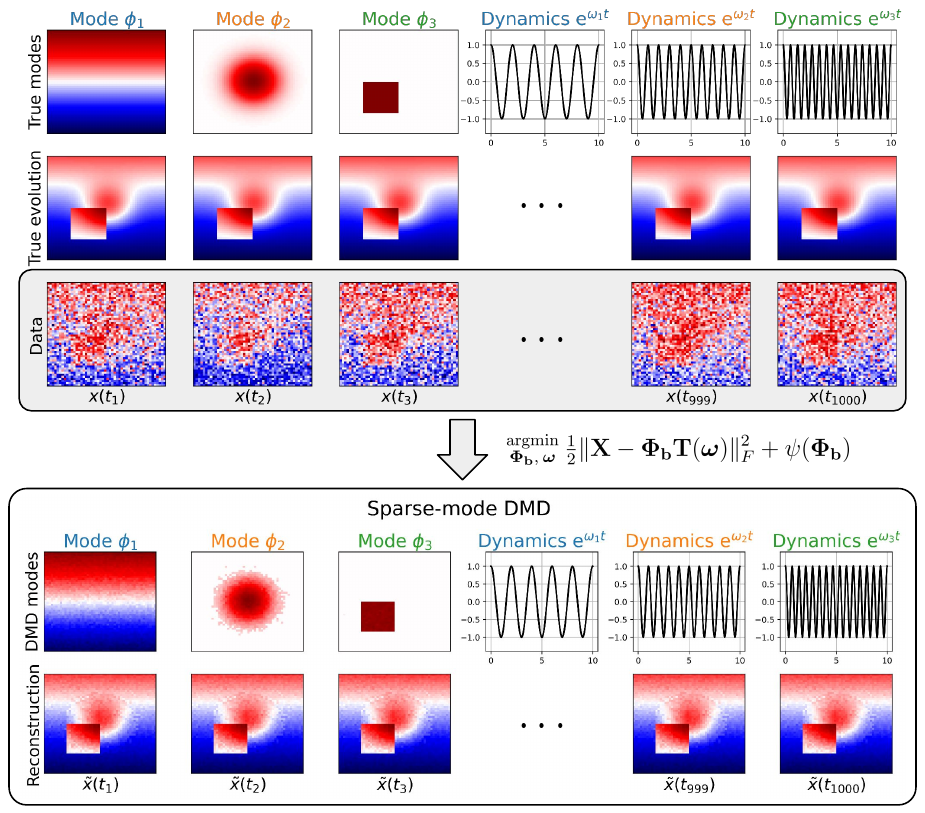}
    \caption{Data set analyzed in Example 4.1. Each data snapshot is a $50 \times 50$ pixel video frame, with the underlying spatiotemporal modes of the system being a slow oscillating gradient, an oscillating Gaussian, and fast oscillating square. The data set, which is corrupted with high levels of Gaussian noise ($\sigma = 0.8$), is fed into the sparse-mode DMD algorithm. By utilizing variable projection and enforcing sparsity in the DMD modes, we are able to robustly recover clean, nearly noise-free spatial modes, along with the true underlying time dynamics associated with each mode.}
    \label{fig:example-1-data}
\end{figure}

In order to demonstrate the general effects of sparsity on the DMD pipeline, we first examine a synthetic video data set that consists of three spatiotemporal modes. The first mode is a gradient that spans the entire video frame and oscillates at the frequency $f_1 = 0.5$ cycles per second, or $\omega_1 = \pi i$. The second mode is a Gaussian at the center of the video, which oscillates at a faster frequency of $f_2 = 1$ cycle per second, or $\omega_2 = 2 \pi i$. Finally, the third mode is a small square at the bottom corner of the video frame, which oscillates the fastest with $\omega_3 = 3 \pi i$. Each frame of the video is $50 \times 50$ pixels and we collect $1,000$ frames total, with each frame being separated by a uniform time step of $\Delta t = 0.01$ seconds. Hence $\mathbf{X} \in \mathbb{C}^{2,500 \times 1,000}$ for this particular data set, assuming that we flatten each frame of our data. We additionally pollute the video with Gaussian noise of magnitude $\sigma = 0.8$ for added obscurity in the dynamics. For a visualization of the data set and the true underlying spatiotemporal modes, see Figure \ref{fig:example-1-data}. 

\begin{figure}
    \centering
    \includegraphics[width=\textwidth]{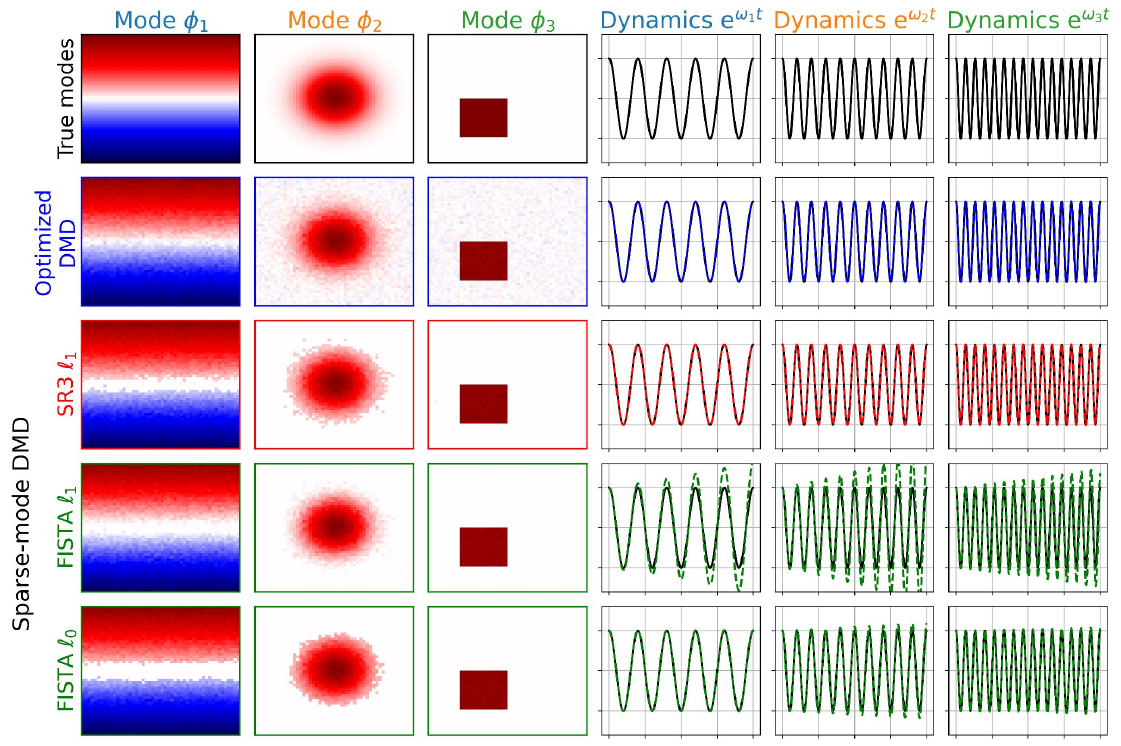}
    \caption{Spatial modes and time dynamics recovered using optimized DMD, sparse-mode DMD with SR3 and $\ell_1$ norm regularization ($\lambda = 0.1$, $\eta = 1$), sparse-mode DMD with FISTA and $\ell_1$ norm regularization ($\lambda = 100$), and sparse-mode DMD with FISTA and $\ell_0$ norm regularization ($\lambda = 10$) to model the data described in Figure \ref{fig:example-1-data}.}
    \label{fig:example-1-results}
\end{figure}

We apply optimized DMD, sparse-mode DMD with FISTA, and sparse-mode DMD with SR3 to this data set using rank $r = 3$. The resulting spatial modes and time dynamics are shown in Figure \ref{fig:example-1-results}. In general, we see that all methods are able to successfully identify the three underlying spatial modes of the system. However due to the high level of noise pollution in the data, optimized DMD produces a dense representation of the Gaussian and square mode, while the sparse methods are able to clear out this residual noise.

With FISTA, the choice of regularizer impacts not only the resulting spatial modes, but also the resulting eigenvalues. $\ell_1$ norm regularization results in modes with softer edges, as should be expected given its proximal operator, however it also introduces a small amount of bias in the resulting eigenvalues. By contrast, $\ell_0$ norm regularization results in modes with rigid edges and very little eigenvalue bias. If we focus on the resulting time dynamics for FISTA with $\ell_1$ in Figure \ref{fig:example-1-results}, we see that the eigenvalues are specifically biased to have an excessively large real component in this example, as evidenced by the exponential growth. However, the frequencies given by the imaginary components of the eigenvalues are computed correctly, hence correction via eigenvalue projection onto the imaginary axis largely mitigates this issue, at least for this particular example. By contrast, the SR3 approach to sparse-mode DMD appears to balance the positive qualities of the optimized DMD and FISTA approaches, as the method de-noises the modes due to sparsity promotion, and it produces eigenvalues with minimal bias largely due to the de-biasing step. The method also converges very quickly, as it requires the least outer variable projection iterations and inner mode matrix updates across all sparse-mode DMD methods examined. This information is highlighted in Table \ref{tab:runtime}. Although this reduction in iterations is somewhat minimal due to the size of the toy data set, we will see that this difference becomes much more pronounced for extremely high-dimensional data sets like the sea surface temperature data examined in Section \ref{sec:ex3}.

As a final experiment, we examine the impact of sparsity strength on model accuracy, both with and without global-local mode separation. To do this, we apply sparse-mode DMD with SR3 and $\ell_1$ norm regularization of varying intensity $\lambda$ to the noisy data set and examine how this impacts our ability to reconstruct the noise-free data. Here we use a rank $r=4$ model in order to highlight the effect of sparsity on low-amplitude, noise-dominated modes. Note that since this particular system only consists of three spatiotemporal modes, the fourth mode is an irrelevant mode that purely consists of noise.

We see from the results in Figure \ref{fig:example-1-sparsity} that for sufficiently low sparsity levels, the modes produced by sparse-mode DMD are the same as those produced by optimized DMD, which is to be expected. Then as we increase our sparsity parameter $\lambda$, we begin to clear out the noise in our modes and our reconstruction error decreases as a result. Whether we use the global-local split or not, there is a certain point at which we sparsify too much. In particular, we find that at a certain point, we begin to sacrifice too much of the Gaussian mode and we begin to essentially trade model accuracy for model sparsity. However, if we sparsify all of our modes, i.e. if we do not perform a global-local split, we find that we reach this point faster due to the over-sparsification of not only the Gaussian mode, but also the gradient mode, which is spatially global by construction. By contrast, if we deploy the global-local split, we are able to correctly identify the gradient as a global mode and apply slightly more sparsity for more gains in model accuracy.

\begin{figure}
    \centering
    \includegraphics[width=\textwidth]{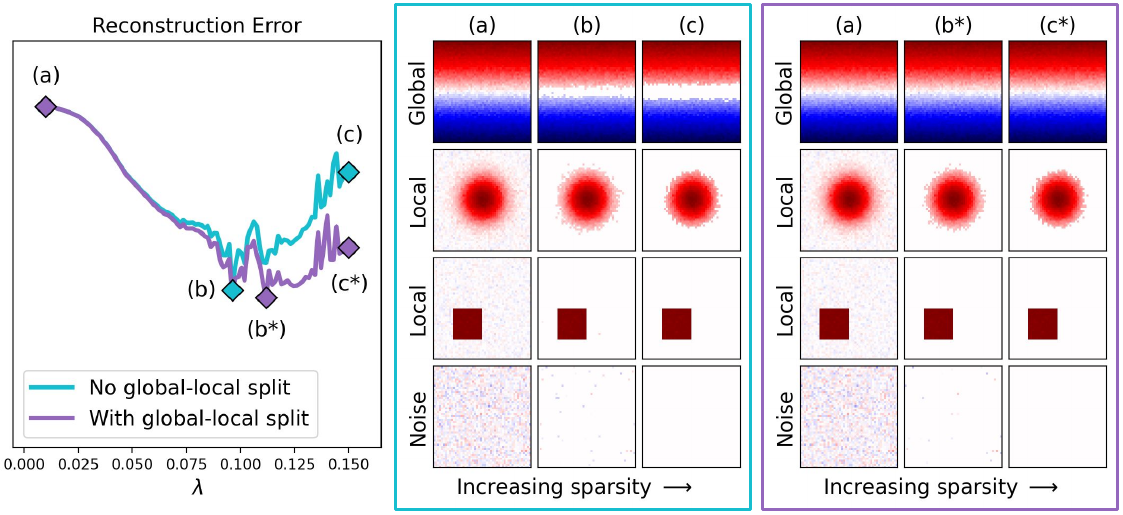}
    \caption{Sparse-mode DMD modes and reconstruction error as a function of model sparsity, both with and without global-local mode splitting. Here we use the sparse-mode DMD with SR3 and $\ell_1$ norm regularization.}
    \label{fig:example-1-sparsity}
\end{figure}

\subsection{Optical waveguides}
\label{sec:ex2}

For our next example, we examine the evolution of an electric field through optical waveguides, whose governing partial differential equation is generally given by the following \cite{kutz_568}:
\begin{equation}
    i u_t + \tfrac{1}{2} u_{xx} - n(x, t) u = \epsilon F(u, x, t).
\label{eq:waveguides}
\end{equation}
$u(x, t)$ denotes the normalized electric field, $n(x, t)$ is a normalized index of refraction profile that generates the waveguide, and $F(u, x, t)$ defines any external forcing or nonlinearities that act upon the system. The parameter $\epsilon$ thus controls the strength of the forcing or nonlinearities.

\subsubsection{The time-dependent Schr\"{o}dinger equation with a square well potential}
In this example, we consider the case where $\epsilon = 0$ and there is hence no forcing or nonlinearities acting upon the system. We also define $n(x, t)$ to be the time-independent potential well
\begin{equation}
    n_{0}(x) = 
    \begin{cases}
        -V_0 & |x| < a \\ 
        \:\;\; 0 & \text{else,}
    \end{cases}
\label{eq:n0}
\end{equation}
specifically for $V_0 = 1$ and $a = 5$. This makes it so that our governing equation of interest is now
\begin{equation}
    i u_t + \tfrac{1}{2} u_{xx} - n_0(x) u = 0,
\label{eq:td_schrodinger}
\end{equation}
which is simply the time-dependent Schr\"{o}dinger equation with a square well potential \cite{bransden2000, griffiths2018}. Using separation of variables and the leading-order solution
\begin{equation}
    u(x, t) = v(x)e^{-i \lambda t}
\end{equation}
for $\lambda \in \R$, we note that the ground-truth solution of \eqref{eq:td_schrodinger} is given by the linear superposition
\begin{equation}
    u(x, t) = \sum_{j=1}^\infty b_j v_j(x) e^{-i \lambda_j t},
\label{eq:true_u}
\end{equation}
where $v_j$ denote the eigenfunctions of the Hamiltonian operator $H = -(1/2) \nabla^2 + n_0(x)$, $\lambda_j$ denote the corresponding eigenvalues, and $b_j$ denote the amplitudes of the spatiotemporal components. In the field of quantum mechanics, the $j$th eigenfunction eigenvalue pair describes a solution state with energy $\lambda_j$, where the spatial component $v_j$ solves the time-independent Schr\"{o}dinger equation
\begin{equation}
    -\tfrac{1}{2} v_j''(x) + n_0(x) v_j(x) = \lambda_j v_j(x).
\label{eq:ti_schrodinger}
\end{equation}

In order to obtain the ground truth eigenfunctions $v_j$, one can solve \eqref{eq:ti_schrodinger} directly, which has well-known solutions for the square potential well \cite{bransden2000, griffiths2018}. In general, we find that
\begin{equation}
    v_j(x) =
    \begin{cases} 
        A e^{\beta_j x} + B e^{-\beta_j x} & x < -a \\
        C \sin(\alpha_j x) + D \cos(\alpha_j x) & |x| < a \\ 
        E e^{\beta_j x} + F e^{-\beta_j x} & x > a
    \end{cases}
\label{eq:true_v}
\end{equation}
for constants $A, B, C, D, E, F \in \C$, where we define the quantities
\begin{align}
    \alpha_j &= \sqrt{2(V_0 + \lambda_j)} &
    \beta_j &= \sqrt{-2 \lambda_j}.
\end{align}
Note that since $-V_0 < \lambda_j$, as the energy of any single solution state cannot exceed the energy of the potential well, $\alpha_j$ is always real. States where $\lambda_j < 0$ are called ``bound states" and make up the discrete spectrum of the Hamiltonian operator. There is a finite number of such states, and they can be defined as either even or odd functions given by the following. Note that several constants from \eqref{eq:true_v} are set to zero in order to prevent infinite growth as $x \to \pm \infty$.
\begin{align}
    v_{j, \, \text{even}}^{d}(x) &=
    \begin{cases} 
        B e^{\beta_j x} & x < -a \\ 
        D \cos(\alpha_j x) & |x| < a \\ 
        B e^{-\beta_j x} & x > a
    \end{cases} & 
    v_{j, \, \text{odd}}^{d}(x) &=
    \begin{cases}
        -A e^{\beta_j x} & x < -a \\ 
        C \sin(\alpha_j x) & |x| < a \\ 
        A e^{-\beta_j x} & x > a
    \end{cases}
\label{eq:true_v_d}
\end{align}
Enforcing continuity at $x=\pm a$ for both $v_j(x)$ and $v'_j(x)$ requires that $\alpha_j\tan(\alpha_j a) = \beta_j$ in the even case and that $\alpha_j\cot(\alpha_j a) = -\beta_j$ in the odd case. These conditions can be rewritten as
\begin{align}
    \xi_j \tan(\xi_j) &= \eta_j & 
    \xi_j \cot(\xi_j) &= -\eta_j
\label{eq:v_d_constraint_1}
\end{align}
via the change of variables $\xi_j = \alpha_j a$ and $\eta_j = \beta_j a$. Using the definition of $\alpha_j$ and $\beta_j$, we have
\begin{equation}
    \xi_j^2 + \eta_j^2 = 2 V_0 a^2.
\label{eq:v_d_constraint_2}
\end{equation}
Hence we find that the depth $V_0$ as well as the length $a$ of the potential well both impact the total number of bound states that are induced by $n_0(x)$, as the intersections of \eqref{eq:v_d_constraint_1} and \eqref{eq:v_d_constraint_2} define the energy levels of the bound states. For our specific potential well defined by $V_0 = 1$ and $a = 5$, this results in $D=5$ bound states total. We then relate our constants $A, B, C, D$ via
\begin{align}
    B &= D\cos(\alpha_j a)e^{\beta_j a} &
    A &= C\sin(\alpha_j a)e^{\beta_j a},
\end{align}
which we derive via continuity in $v_j(x)$ at $x = \pm a$. We additionally normalize each $v_j$ such that
\begin{equation}
    \langle v_j, v_j\rangle = \int_{-\infty}^{\infty} \left|v_j(x)\right|^2 dx = 1.
\label{eq:normalization}
\end{equation}
Finally, all of these relationships combined allow us to use \eqref{eq:true_v_d} in order to explicitly compute the ground truth bound states that are induced by the potential well $n_0(x)$, which we visualize in Figure \ref{fig:example-2-modes}.  The analytic derivation presented here is often limited to linear operators, but there are special cases with square wave potentials with nonlinearity which also allow for analytic reduction~\cite{mahmud2002bose}.

States where $\lambda_j > 0$ are alternatively called ``scattering states" and make up what is known as the continuous spectrum of the Hamiltonian. There is an infinite number of such states, and they can be defined as the following. Note that we set $F = 0$ from \eqref{eq:true_v} due to the assumption that the wave is solely transmitted and not reflected in this section of the domain.
\begin{equation}
    v_j^c(x) =
    \begin{cases}
        A e^{\beta_j x} + B e^{-\beta_j x} & x < -a \\ 
        C \sin(\alpha_j x) + D \cos(\alpha_j x) & |x| < a \\ 
        E e^{\beta_j x} & x > a
    \end{cases}
\label{eq:true_v_c}
\end{equation}
Enforcing continuity at $x=\pm a$ for $v_j(x)$ and $v'_j(x)$ in the scattering case yields
\begin{align}
    Ae^{-\beta_j a} + Be^{\beta_j a} &= -C \sin(\alpha_j a) + D \cos(\alpha_j a) \\ 
    \beta_j \big[ A e^{-\beta_j a} - B e^{\beta_j a} \big] &= \alpha_j \big[ C \cos(\alpha_j a) + D \sin(\alpha_j a) \big] \\ 
    E e^{\beta_j a} &= C \sin(\alpha_j a) + D \cos(\alpha_j a) \\ 
    \beta_j E e^{\beta_j a} &= \alpha_j \big[ C \cos(\alpha_j a) - D \sin(\alpha_j a) \big],
\end{align}
which results in the following relationships between the constants $A, B, C, D, E$:
\begin{align}
    A &= E \left[ \cos(2 \alpha_j a ) + \left( \frac{\alpha_j^2 - \beta_j^2}{2 \alpha_j \beta_j} \right) \sin(2 \alpha_j a) \right] e^{2 \beta_j a} & 
    C &= E \left[ \sin(\alpha_j a) + \tfrac{\beta_j}{\alpha_j} \cos(\alpha_j a) \right] e^{\beta_j a} \\ 
    B &= -E \left( \frac{\alpha_j^2 + \beta_j^2}{2\alpha_j \beta_j} \right) \sin(2 \alpha_j a) &
    D &= E \left[ \cos(\alpha_j a) - \tfrac{\beta_j}{\alpha_j} \sin(\alpha_j a) \right] e^{\beta_j a}.
\end{align}
Additionally enforcing normalization \eqref{eq:normalization} then allows us to use these relationships in conjunction with \eqref{eq:true_v_c} in order to define the ground truth scattering states for a given $\lambda_j$, much like in the bound case.

\begin{figure}
    \centering
    \includegraphics[width=\textwidth]{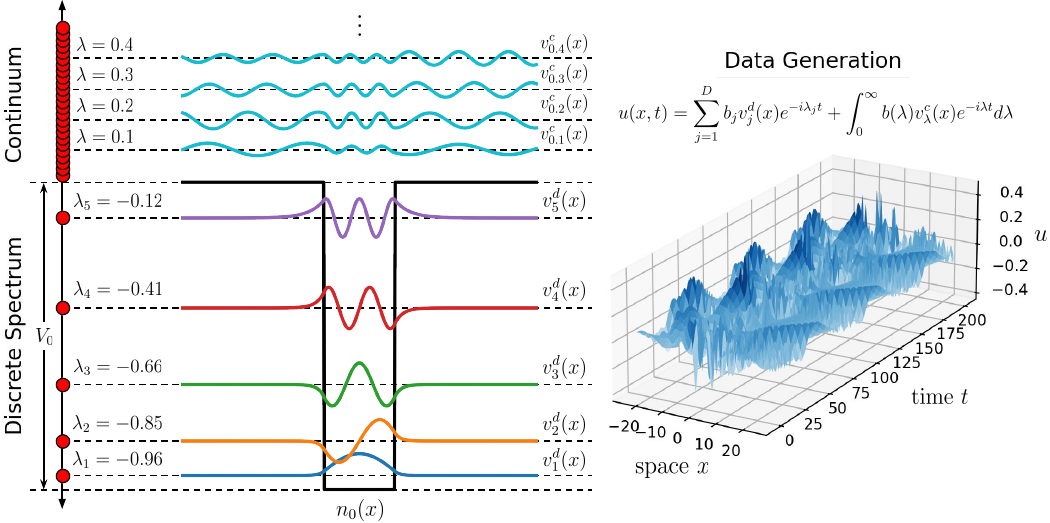}
    \caption{Eigenfunctions $v_j(x)$ and their corresponding eigenvalues $\lambda_j$ of the Hamiltonian operator of the time-independent Schr\"{o}dinger equation \eqref{eq:ti_schrodinger}. The Hamiltonian possesses $D=5$ eigenfunctions with $\lambda_j < 0$, whose activities are bound to the potential well $n_0(x)$. The Hamiltonian additionally possesses infinitely-many eigenfunctions with $\lambda > 0$, whose activities are spatially global and not bound to the potential well. Combining of these spatiotemporal components via linear superposition \eqref{eq:true_u2} gives rise to complex time-varying data sets. The particular data set featured here was generated using all components of the discrete spectrum with amplitude $b = 0.5$, and the components of the continuous spectrum for $\lambda = 0.1, 0.2, \dots, 49.9, 50$ with amplitudes $b = 0.2/\lambda$.
    }
    \label{fig:example-2-modes}
\end{figure}

Unlike for the bound states \eqref{eq:true_v_d}, the continuity constraints for the scattering states \eqref{eq:true_v_c} can be achieved with any $\lambda_j > 0$, making it so that all energy levels $\lambda_j > 0$ are eigenvalues of the Hamiltonian. This means that \eqref{eq:true_u} can alternatively be expressed via two components: a discrete summation component that captures the behavior of the bound states, and a continuous integral component that captures the behavior of the scattering states.
\begin{equation}
    u(x, t) = \sum_{j=1}^D b_j v_j^d(x) e^{-i \lambda_j t} + \int_{0}^{\infty} b(\lambda) v^c_\lambda(x) e^{-i \lambda t} d\lambda
\label{eq:true_u2}
\end{equation}

In order to simulate this system, we define the snapshot vectors $U(t) \in \R^n$, which approximate the electric field $u(x, t)$ at the spatial collocation points $x_1, x_2, \dots, x_n$ at time $t$. For our particular simulation of the data, we examine the spatial grid $x \in [-25, 25]$ at times $t=0, 1, \dots, 199$ using the uniform grid spacing $\Delta x = 0.1$, resulting in 500 collocation points in space and 200 collocation points in time to form $\mathbf{X} \in \R^{500 \times 200}$. We then use our ground truth expressions for the bound states \eqref{eq:true_v_d} and scattering states \eqref{eq:true_v_c}, along with Equation \eqref{eq:true_u2} in order to generate our data.

\begin{figure}
    \centering
    \includegraphics[width=\textwidth]{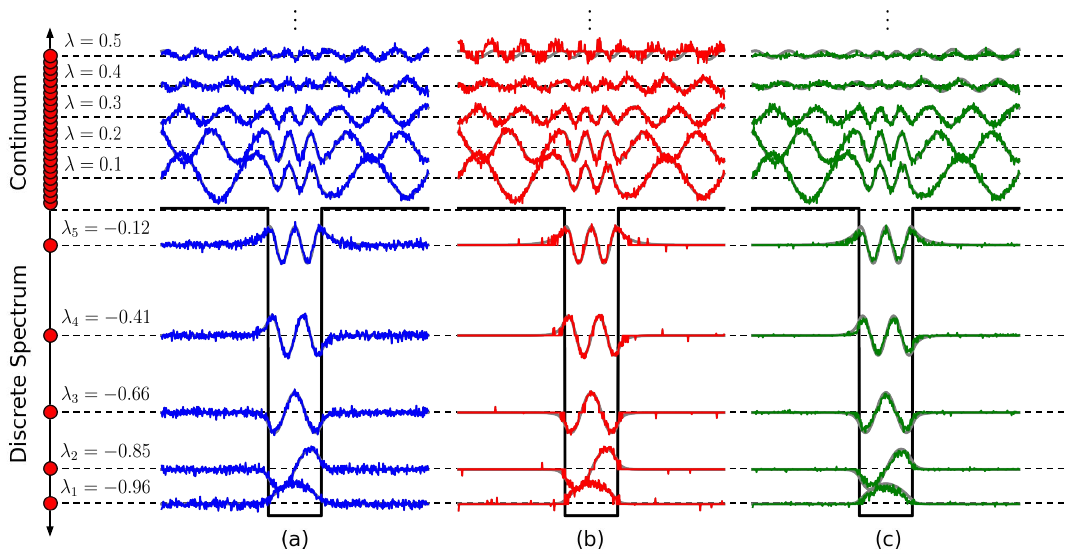}
    \caption{Eigenfunctions and eigenvalues recovered by (a) optimized DMD, (b) sparse-mode DMD with SR3 ($\ell_1$ norm sparsity with intensity $\lambda = 0.02125$), and (c) sparse-mode DMD with FISTA ($\ell_1$ norm sparsity with intensity $\lambda = 2.5$ and projected eigenvalues) when applied to the data set in Figure \ref{fig:example-2-modes}, which has been polluted by Gaussian noise of magnitude $\sigma = 0.15$.}
    \label{fig:example-2-results}
\end{figure}

\subsubsection{Results of applying DMD}

We simulate the optical waveguide system using all $D = 5$ bound states, each with amplitude $b = 0.5$. We additionally use 500 scattering states, which correspond with energy levels $\lambda = 0.1, 0.2, \dots, 49.9, 50$. We assign each scattering state the amplitude $b = 0.2 / \lambda$ so that the prominence of a single scattering state is inversely proportional to its energy level. In doing so, we produce a system that consists of 505 spatiotemporal modes total, with the most prominent modes being the discrete spectrum and the low-energy continuous states. We note that the sheer number of modes present makes the task of computing a finite DMD representation much harder than in the previous example. However in this case, the goal is not to perfectly reconstruct the data, but rather to accurately recover the most prominent spatiotemporal modes of our system from noisy, limited data. As such, we additionally pollute the data with Gaussian noise of magnitude $\sigma = 0.15$ prior to applying DMD.

We apply optimized DMD, sparse-mode DMD with FISTA, and sparse-mode DMD with SR3 to this data set, where we use rank $r = 20$ for each DMD model. See Figure \ref{fig:example-2-results} for a visualization of the resulting modes and eigenvalues for each method. Note that both sparse-mode DMD methods are applied using $\ell_1$ norm regularization, with the FISTA approach utilizing eigenvalue projection onto the imaginary axis in accordance with the observations from the previous example. From these results, we see that both optimized DMD and sparse-mode DMD are able to identify the modes of the discrete spectrum, as well as the most prominent modes of the continuous spectrum. The modes extracted by both methods contain inaccuracies due to the presence of noise and the general difficulty of representing this system with a discrete set of modes. However, we see that sparse-mode DMD is able to correctly and automatically eliminate the extraneous activities detected in the modes of the discrete spectrum. Additionally, we see that with high enough levels of sparsity, sparse-mode DMD begins to eliminate the lowest-amplitude modes while still preserving dominant global and local features.

We again observe small differences in the FISTA and SR3 approaches to sparse-mode DMD. In particular, FISTA with $\ell_1$ norm sparsity yields spatially smoother modes, while SR3 with $\ell_1$ norm sparsity yields modes with harsh edges but slightly more accurate peaks, similar to what was observed in the previous example. Both methods correctly identify the 3 lowest-energy scattering states as prominent global modes that should not be sparsified, while sparsifying the remaining modes which are characterized by either spatially local features or small amplitudes. It is worth noting however that toggling the parameters of the global-local splitting procedure will impact which modes are identified as global modes. We also observe once again based on the results of Table \ref{tab:runtime} that sparse-mode DMD with SR3 converges in fewer iterations and overall less time than sparse-mode DMD with FISTA.

\subsection{Sea surface temperature data}
\label{sec:ex3}

\begin{figure}
    \centering
    \includegraphics[width=\textwidth]{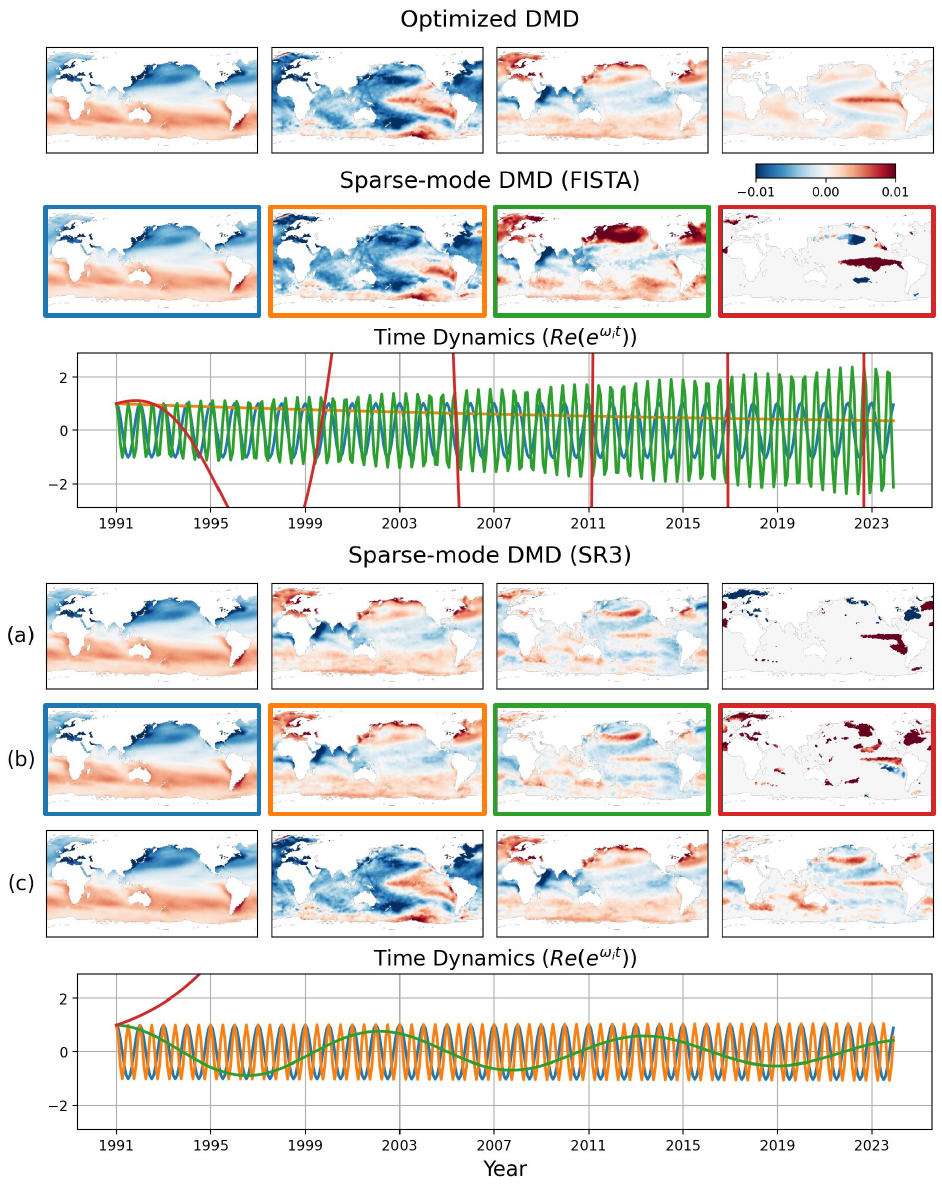}
    \caption{The 4 leading DMD modes discovered by optimized DMD, sparse-mode DMD with FISTA ($\ell_0$ norm sparsity with intensity $\lambda = 31,622$), and (a-c) sparse-mode DMD with SR3 ($\ell_1$ norm sparsity with intensities $\lambda = 0.007743, 0.02069, 2.336$) when applied to the HadISST data set \cite{raynerGlobalAnalysesSea2003}. Corresponding time dynamics are provided for the FISTA model and for SR3 model (b).}
    \label{fig:example-3-results}
\end{figure}

As our final example for demonstrating sparse-mode DMD, we examine sea surface temperature data. Specifically, we examine the Hadley Centre Sea Ice and Sea Surface Temperature (HadISST) data set \cite{raynerGlobalAnalysesSea2003}, which contains monthly sea surface temperature readings from across the globe from the year 1991 to 2023\footnote{Data publicly available at \href{https://www.metoffice.gov.uk/hadobs/hadisst/}{https://www.metoffice.gov.uk/hadobs/hadisst/}.}. The data consists of $396$ snapshots, where each snapshot records sea surface temperatures on a $360 \times 180$ spatial grid. Omitting grid points that account for land or sea-ice and flattening each snapshot results in $34,504$-dimensional snapshots, making it so that $\mathbf{X} \in \R^{34,504 \times 396}$. Sea surface temperature data is known to contain multiscale phenomena, which includes El Niño and La Niña warming events that cannot be modeled using simple DMD models \cite{dmd_book, mrcosts}. Hence much like in the case of the previous example, our primary goal is not to perfectly model the data with DMD, but rather to use DMD as a tool for unveiling dominant global and local spatiotemporal features that may be embedded within the data.

The spatial modes and time dynamics that result from applying optimized DMD and sparse-mode DMD with rank $r=8$ are visualized in Figure \ref{fig:example-3-results}. Note that for this particular data set, we default to sparse-mode DMD with SR3, as the computational costs associated with running FISTA become prohibitively expensive due to the size of the data. See Table \ref{tab:runtime} for more details. We also find that sparse-mode DMD with FISTA, though it learns different spatiotemporal features, tends to produce models that are less accurate when compared to the models learned using SR3. This paired with the huge computational costs associated with using FISTA make the FISTA approach unideal for this application. Nonetheless, we provide an example FISTA result in Figure \ref{fig:example-3-results} in order to illustrate the differences that arise between sparse-mode DMD methods.

Based on the results in Figure \ref{fig:example-3-results}, we generally see that by introducing sparsity, we begin to extract and isolate the El Niño warming band via the fourth most prominent DMD mode, as opposed to optimized DMD, which exclusively produces spatially global modes. In addition to this, we again see that sparse-mode DMD is able to learn dominant spatially global modes in addition to these prominent local features. In particular, we see that even as we introduce sparsification, we consistently recover the same leading spatiotemporal mode, which we note is associated seasonal temperature changes that impact the entire globe. Notice that this can be inferred from the corresponding time dynamics, as this phenomenon finishes one cycle each year.

Although the implications of these results are generally unknown and hence require investigation beyond the scope of the current work, DMD's retrieval of global and local features that correspond with learned temporal dynamics raises a variety of interesting questions and potential avenues of exploration. For example, is it possible to treat our learned global modes as members of a ``continuum" and our learned local modes as members of the ``discrete spectrum" and if so, how could we potentially leverage such insight? For many systems, including the optical waveguide system discussed in the previous section, it is possible to induce modal coupling and the exchange of energy between members of the discrete spectrum by forcing the system at a frequency that relies on the frequencies of the coupled modes \cite{kutz_568}. Hence one might explore the temporal frequencies associated with phenomena that force the SST system, such as solar radiation or atmospheric conditions, and examine their relationship (or lack thereof) with the El Niño frequencies learned by sparse-mode DMD in order to better understand modal energy exchange in the context of sea surface temperatures.

\begin{table}[t]
    \centering
    \begin{tabular}{crccll} \toprule
        \makecell{Data set\\$(n,m,r)$} & \makecell{Sparse-mode DMD Method} & \makecell{\# VP\\iter.} & \makecell{\# mode iter. \\ (avg.)} & \makecell{Time \\ (sec.)} & \makecell{Acc.} \\ \toprule
        \multirow{3}{*}{\makecell{Synthetic\\$(2500, 1000, 3)$}}
        & SR3 ($\ell_1, \lambda=10^{-1}$) & 6 & 4.0 & $\tau_{\text{SR3}}$ & 7.298\% \\
        & FISTA ($\ell_1, \lambda=10^{2}$) & 6 & 10.33 & $\tau_{\text{SR3}} \times 2.117$ & 20.18\% \\ 
        & FISTA ($\ell_0, \lambda=10^{1}$) & 24 & 6.833 & $\tau_{\text{SR3}} \times 2.440$ & 8.987\% \\ \toprule
        \multirow{2}{*}{\makecell{Waveguides\\$(500, 200, 20)$}}
        & SR3 ($\ell_1, \lambda=2.125 \times 10^{-2}$) & 8 & 9.375 & $\tau_{\text{SR3}}$ & 25.89\% \\
        & FISTA ($\ell_1, \lambda=2.5 \times 10^0$) & 33 & 39.33 & $\tau_{\text{SR3}} \times 4.268 $ & 26.41\% \\ \toprule
        \multirow{4}{*}{\makecell{SST\\$(34504, 396, 8)$}}
        & SR3 ($\ell_1, \lambda=7.743 \times 10^{-3}$) & 13 & 6.308 & $\tau_0$ & 23.29\% \\ 
        & SR3 ($\ell_1, \lambda=2.069 \times 10^{-2}$) & 15 & 6.4 & $\tau_0 \times 1.770$ & 23.71\% \\ 
        & SR3 ($\ell_1, \lambda=2.336 \times 10^0$) & 14 & 7.643 & $\tau_0 \times 2.266$ & 23.39\% \\
        & FISTA ($\ell_0, \lambda= 3.162 \times 10^{4}$) & 17 & 603.5 & $\tau_0 \times 25.00$ & 29.77\% \\ \toprule
    \end{tabular}
    \caption{Computational costs associated with the sparse-mode DMD methods highlighted in Section \ref{sec:examples}. Outer variable projection iterations were run until convergence of the objective function to tolerance $10^{-6}$, and inner mode update iterations, which occur at every variable projection iteration, were run until convergence of the mode matrix to tolerance $10^{-6}$. The only exception to this was the FISTA model for the SST data set, which used the mode matrix tolerance $10^{-3}$ for improved runtimes. Runtimes were obtained via the Google Colab CPU. Accuracies are given by the relative error in the reconstruction of the clean data set (if available) across all time points. All corresponding code is available at \href{https://github.com/sichinaga/sparse_mode_dmd}{https://github.com/sichinaga/sparse\_mode\_dmd}.}
    \label{tab:runtime}
\end{table}

\section{Discussion}
\label{sec:discussion}
In this work, we introduced a variant of the optimized DMD framework that applies sparsity-promoting regularization to the DMD modes directly in order to promote spatial localization to the relevant modes. We formulated our algorithm using iterative variable projection updates, with Levenberg-Marquardt for updating the DMD eigenvalues, and FISTA or SR3 for updating the mode matrices. We showed that by invoking sparsity on the modes computed by DMD, and by identifying candidate global modes that should not be subject to sparsification, one is able to recover mixtures of dominant spatially global and local modes from noisy data and systems that potentially contain an overwhelming amount of underlying spatiotemporal structures. High-dimensional nonlinear systems are ubiquitous throughout the sciences, with many of these systems often possessing a vast continuum of underlying spatiotemporal modes. Sparse-mode DMD hence has great potential to aid in the analysis of such systems by revealing members of the discrete spectrum in addition to the most-prominent members of the continuum.

Due to the general proximal gradient-based framework utilized by sparse-mode DMD, there is a great deal of potential methodological extensions and improvements regarding the algorithm itself. For example, one natural extension would be the deployment of more sophisticated sparsification routines such as group sparsity promotion \cite{Parikh2014}. Group sparsity would be particularly interesting to deploy since it is reasonable to infer that entire spatiotemporal modes or structures should be sparsified as opposed to individual features. Another natural extension would be to optimize and better-automate the global-local splitting procedure. Although the procedure outlined here functions sufficiently-well for the analyses carried out in this work, this process is sensitive to hyperparameter tuning. A more preferred global-local splitting formulation might involve an additional optimization routine, which seeks to find the optimal set of modes to sparsify and the optimal set of modes to not to sparsify, which would then respectively yield the desired local and global modes.

From an application viewpoint, the sparsity-promoting DMD method provides an important innovation for many practical applications such as quantum mechanics, acoustics and electrodynamics.  Specifically, many such physics based systems have a combination of discrete and continuous spectra that play critical roles in the dynamics and observed behavior.  The ability to automatically disambiguate between local and global structures gives a robust way to group contributions to the dynamics from the discrete and continuous spectra.  In practice, many technological applications rely on ones ability to engineer the characteristics of both these contributions.  This can be done in a a fully data-driven way with the sparsity-promoting DMD method advocated here.  Moreover, in the SST example considered, the method identifies truly localized structures that play important roles in physical systems whose governing equations are only partially known, thus highlighting the role of local modes in interactions with global dynamics.

\section*{Acknowledgments}
This work was supported in part by the US National Science Foundation (NSF) AI Institute for Dynamical Systems (dynamicsai.org), grant 2112085.
JNK further acknowledges support from the Air Force Office of Scientific Research (FA9550-24-1-0141).

\bibliographystyle{siam}
\bibliography{refs}


\end{document}